\documentclass[10pt,twocolumn,letterpaper]{article}

\usepackage[pagenumbers]{cvpr} 

\usepackage{graphicx}
\usepackage{amsmath}
\usepackage{amssymb}
\usepackage{booktabs}
\usepackage{enumitem}
\usepackage{multirow}
\usepackage{subcaption}

\usepackage[pagebackref,breaklinks,colorlinks]{hyperref}

\usepackage[capitalize]{cleveref}
\crefname{section}{Sec.}{Secs.}
\Crefname{section}{Section}{Sections}
\Crefname{table}{Table}{Tables}
\crefname{table}{Tab.}{Tabs.}

\def\cvprPaperID{*****} 
\def\confName{CVPR}
\def\confYear{2022}

\begin{document}

\title{Wiener Guided DIP for Unsupervised Blind Image Deconvolution}

\author{Gustav Bredell$^{1*}$\,, Ertunc Erdil$^1$\,,
Bruno Weber$^2$\,,
Ender Konukoglu$^1$\\
{\small$^1$ Department of Information Technology and Electrical Engineering, ETH-Zurich, Zurich, Switzerland}\\
{\small$^2$Institute of Pharmacology and Toxicology, University of Zurich, Zurich, Switzerland}\\
{\tt\small$*${gustav.bredell@vision.ee.ethz.ch}}
}


\maketitle

\begin{abstract}
   Blind deconvolution is an ill-posed problem arising in various fields ranging from microscopy to astronomy. The ill-posed nature of the problem requires adequate priors to arrive to a desirable solution. Recently, it has been shown that deep learning architectures can serve as an image generation prior during unsupervised blind deconvolution optimization, however often exhibiting a performance fluctuation even on a single image. We propose to use Wiener-deconvolution to guide the image generator during optimization by providing it a sharpened version of the blurry image using an auxiliary kernel estimate starting from a Gaussian. We observe that the high-frequency artifacts of deconvolution are reproduced with a delay compared to low-frequency features. In addition, the image generator reproduces low-frequency features of the deconvolved image faster than that of a blurry image. We embed the computational process in a constrained optimization framework and show that the proposed method yields higher stability and performance across multiple datasets. In addition, we provide the code\footnote{\label{note1}\href{https://github.com/gbredell/W_DIP.git}{https://github.com/gbredell/W\_DIP.git}}. 
   
\end{abstract}


\section{Introduction}
\label{sec:intro}

\begin{figure}[ht]
\centering
\captionsetup[subfigure]{labelformat=empty}
     \begin{subfigure}[b]{0.45\linewidth}
         \includegraphics[width=\linewidth]{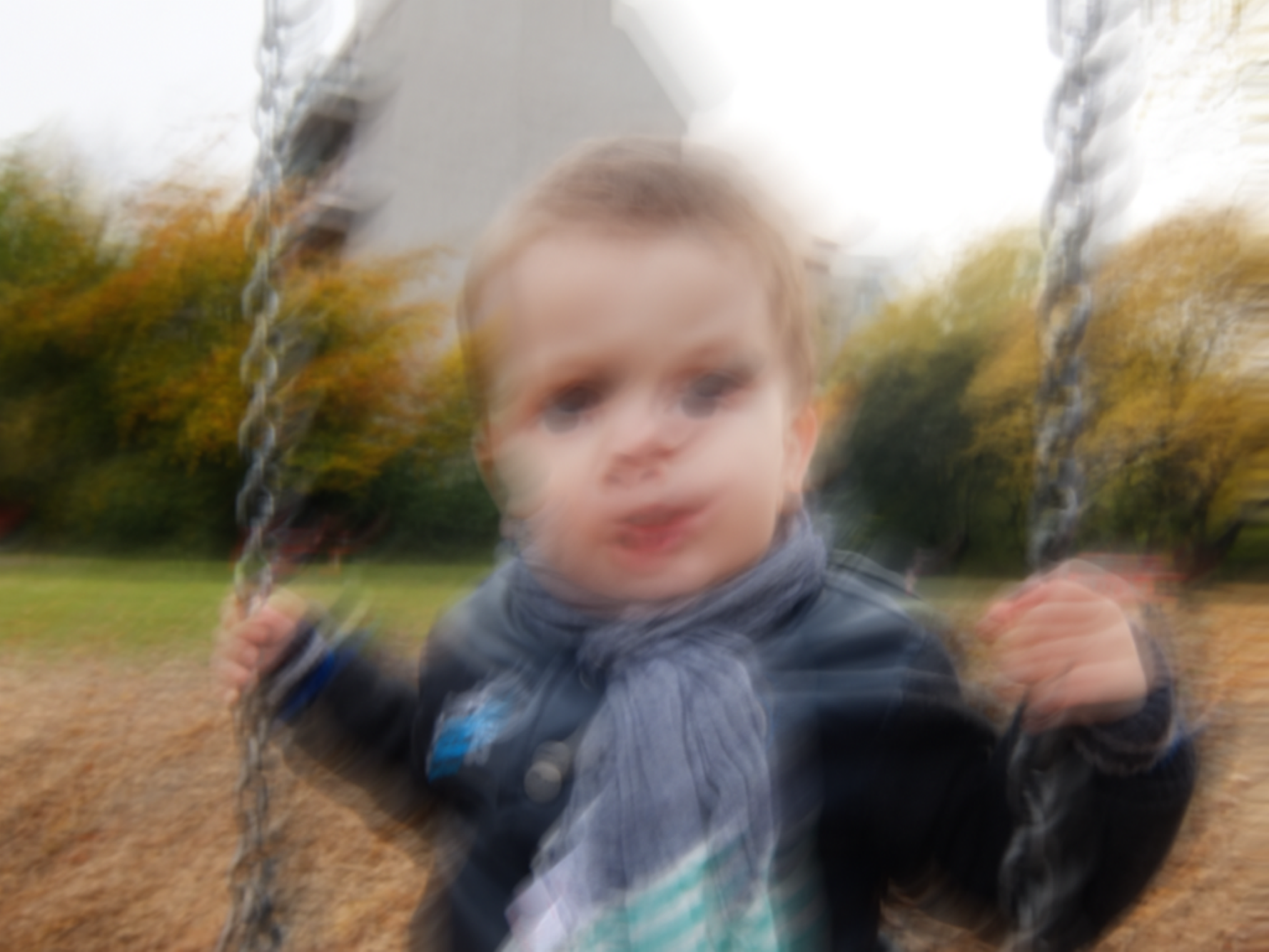}
         \caption{Blurred Image}
         \label{fig:front_blur}
     \end{subfigure}%
     \hspace{0.001\textwidth}
     \begin{subfigure}[b]{0.45\linewidth}
         \includegraphics[width=\linewidth]{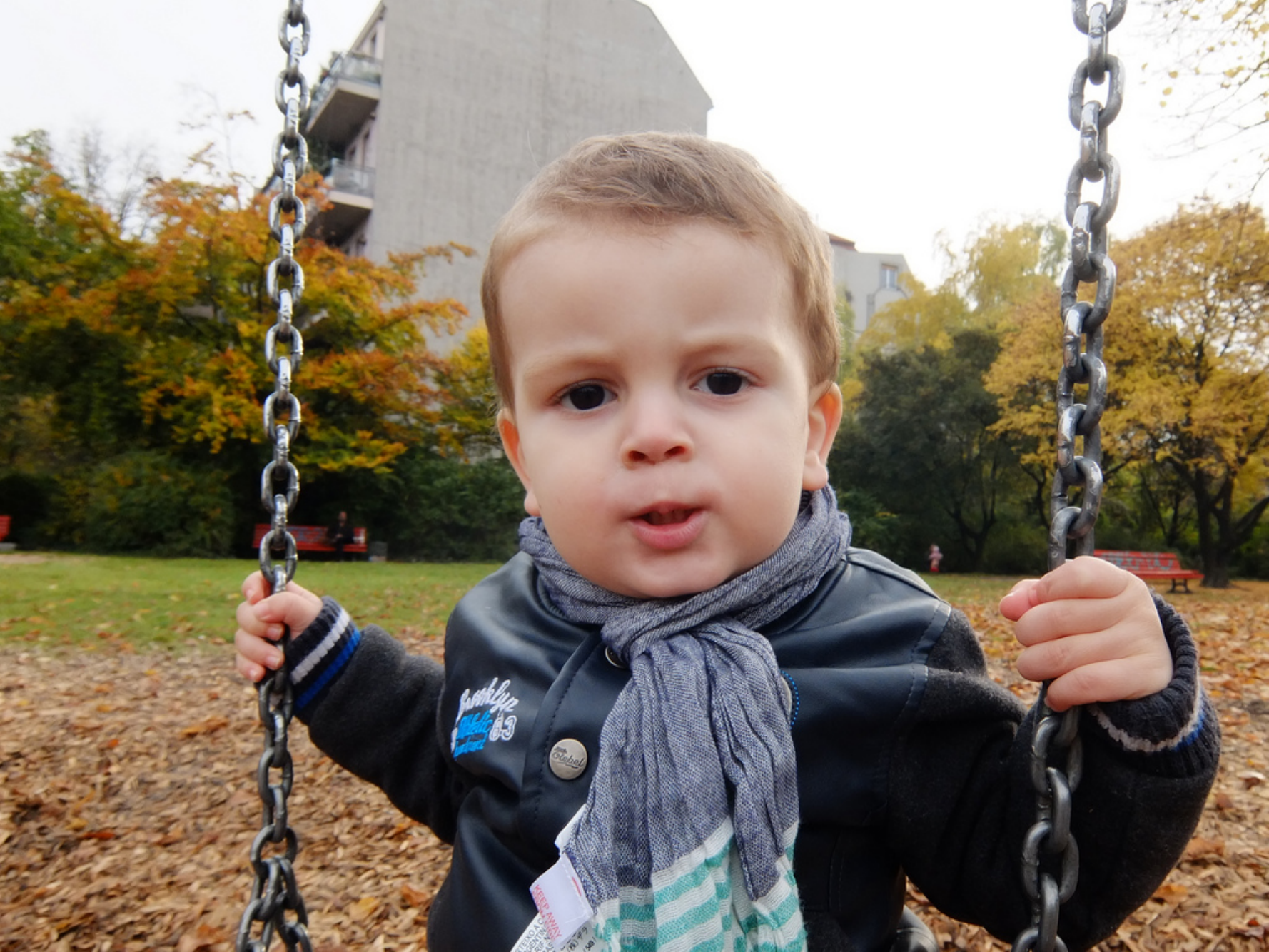}
         \caption{Ground Truth}
         \label{fig:front_gt}
     \end{subfigure}
  \\
     \begin{subfigure}[b]{0.45\linewidth}
         \includegraphics[width=\linewidth]{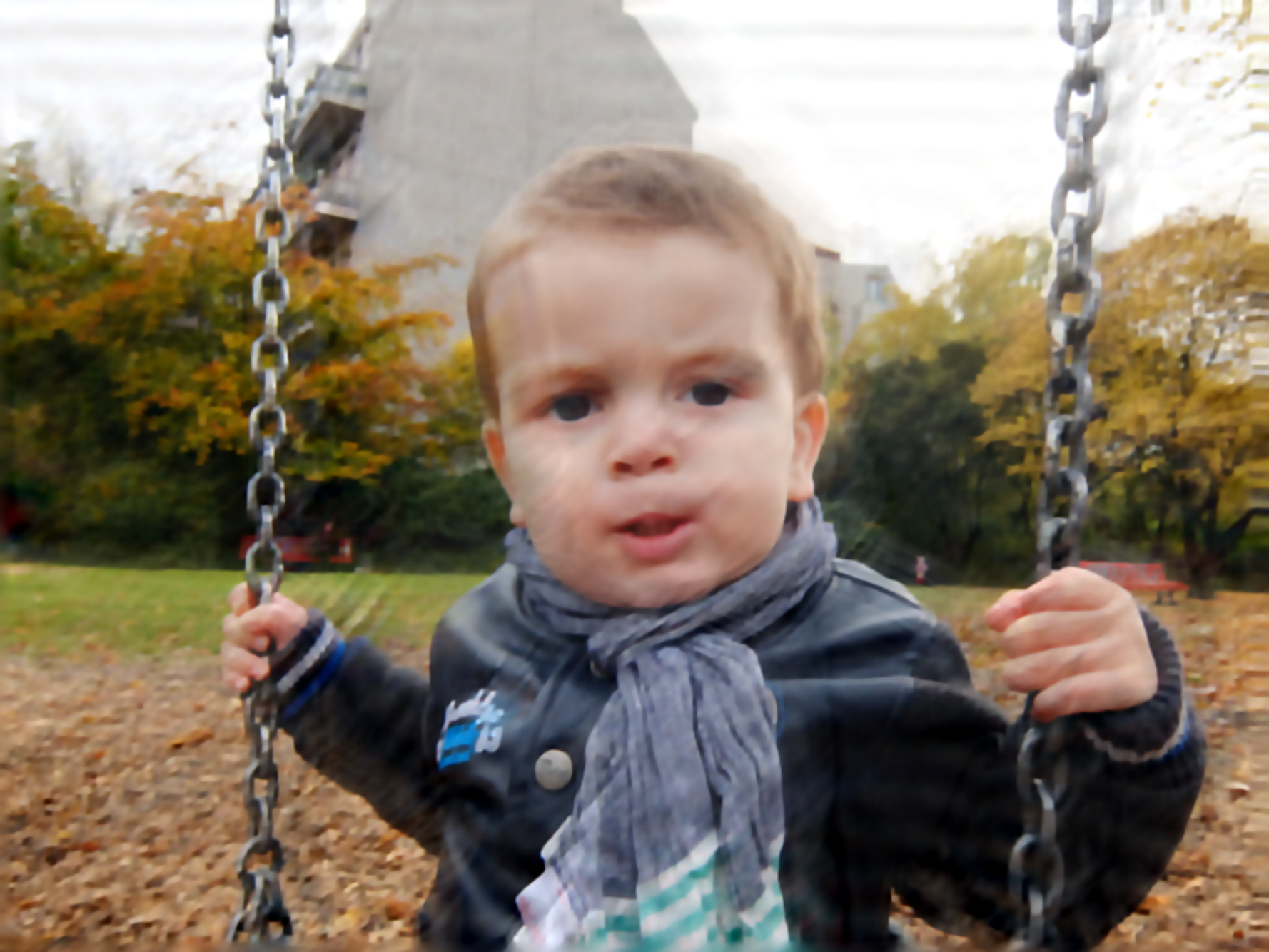}
         \caption{Michaeli \etal~\cite{michaeli2014blind}}
         \label{fig:front_michaeli}
     \end{subfigure}%
     \hspace{0.001\textwidth}
     \begin{subfigure}[b]{0.45\linewidth}
         \includegraphics[width=\linewidth]{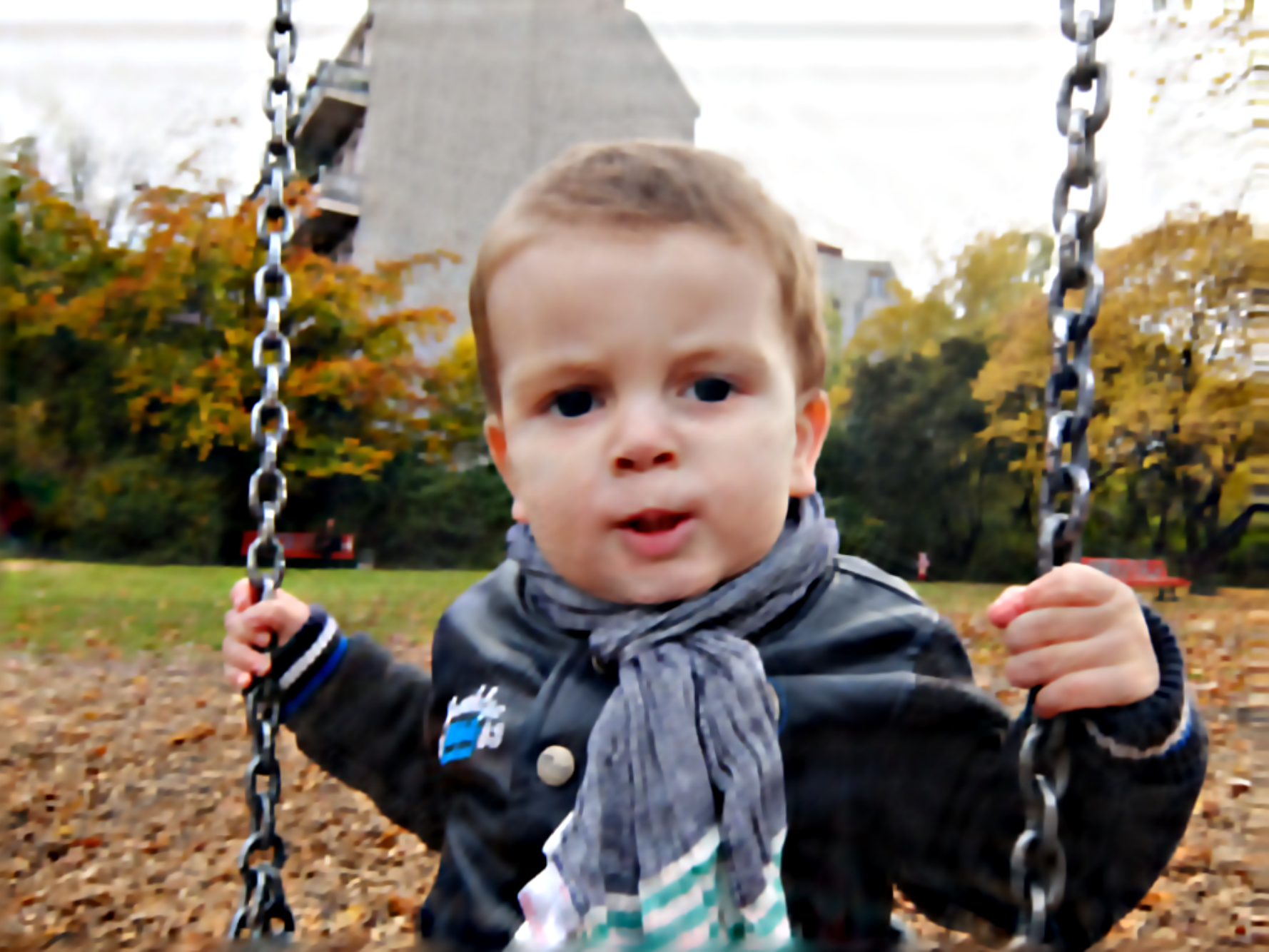}
         \caption{Perrone \etal~\cite{perrone2014total}}
         \label{fig:front_perrone}
     \end{subfigure}
  \\
     \begin{subfigure}[b]{0.45\linewidth}
         \includegraphics[width=\linewidth]{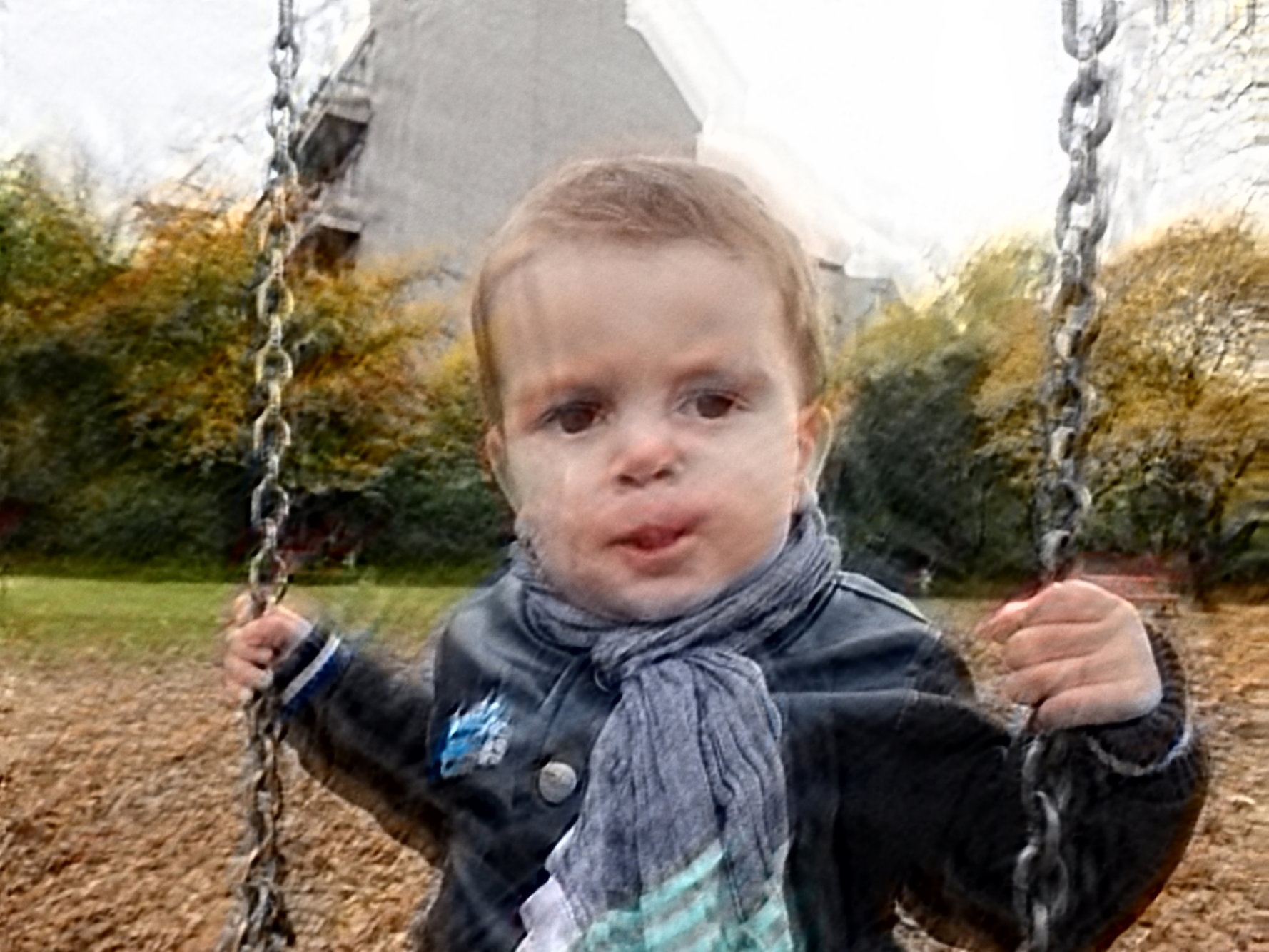}
         \caption{Ren \etal~\cite{ren2020neural}}
         \label{fig:front_SelfDeblur}
     \end{subfigure}%
     \hspace{0.001\textwidth}
     \begin{subfigure}[b]{0.45\linewidth}
         \includegraphics[width=\linewidth]{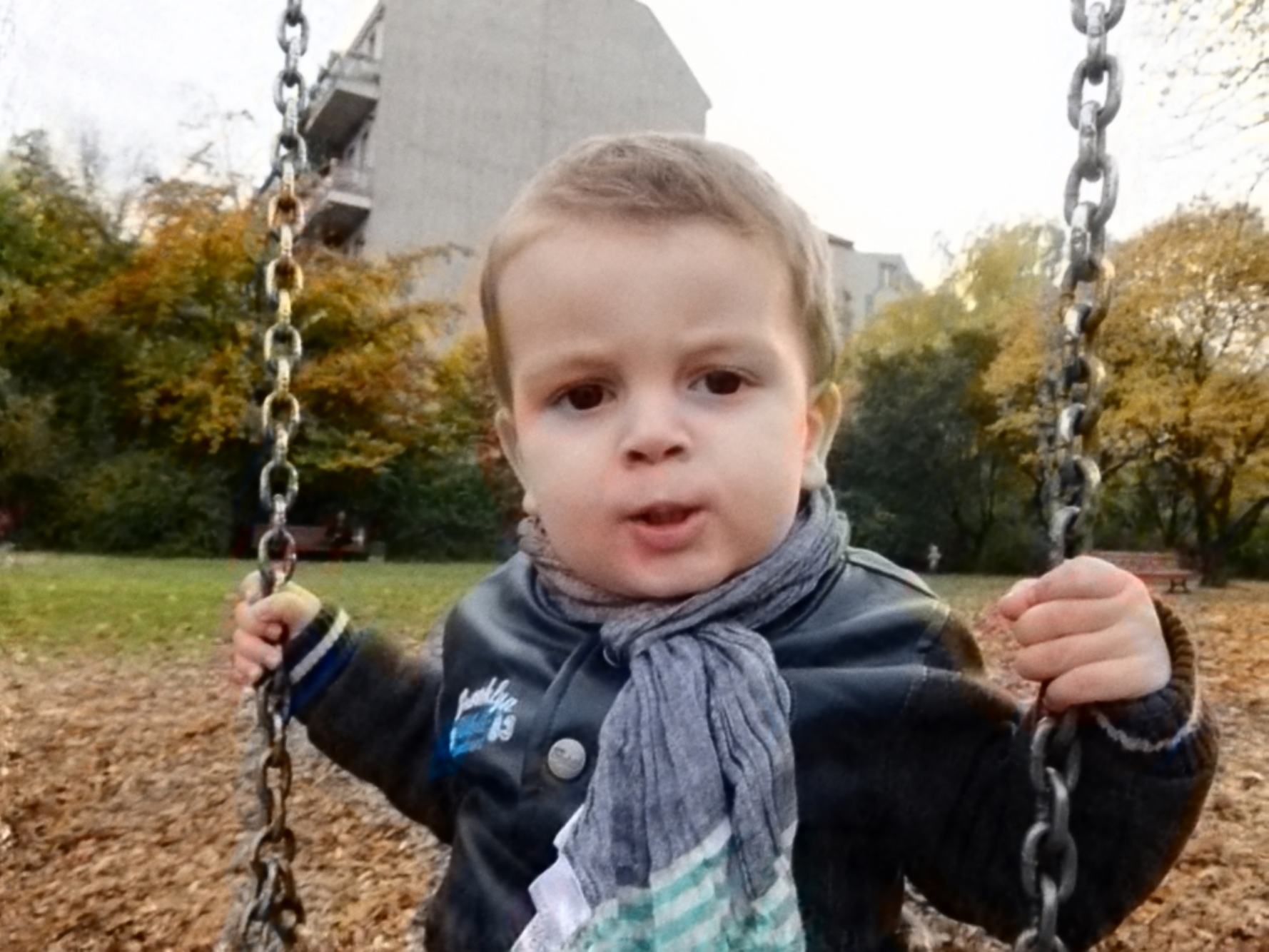}
         \caption{W-DIP}
         \label{fig:front_WDIP}
     \end{subfigure}
     
        \caption{Motivating example showing that while the existing methods produces artifact and blurry results especially around the trees in the background, W-DIP produces the most similar result to the ground truth image.}
        \label{fig:front_figure}
\end{figure}

\begin{figure*}[ht]
     \begin{subfigure}[b]{0.48\linewidth}
         \includegraphics[width=\linewidth]{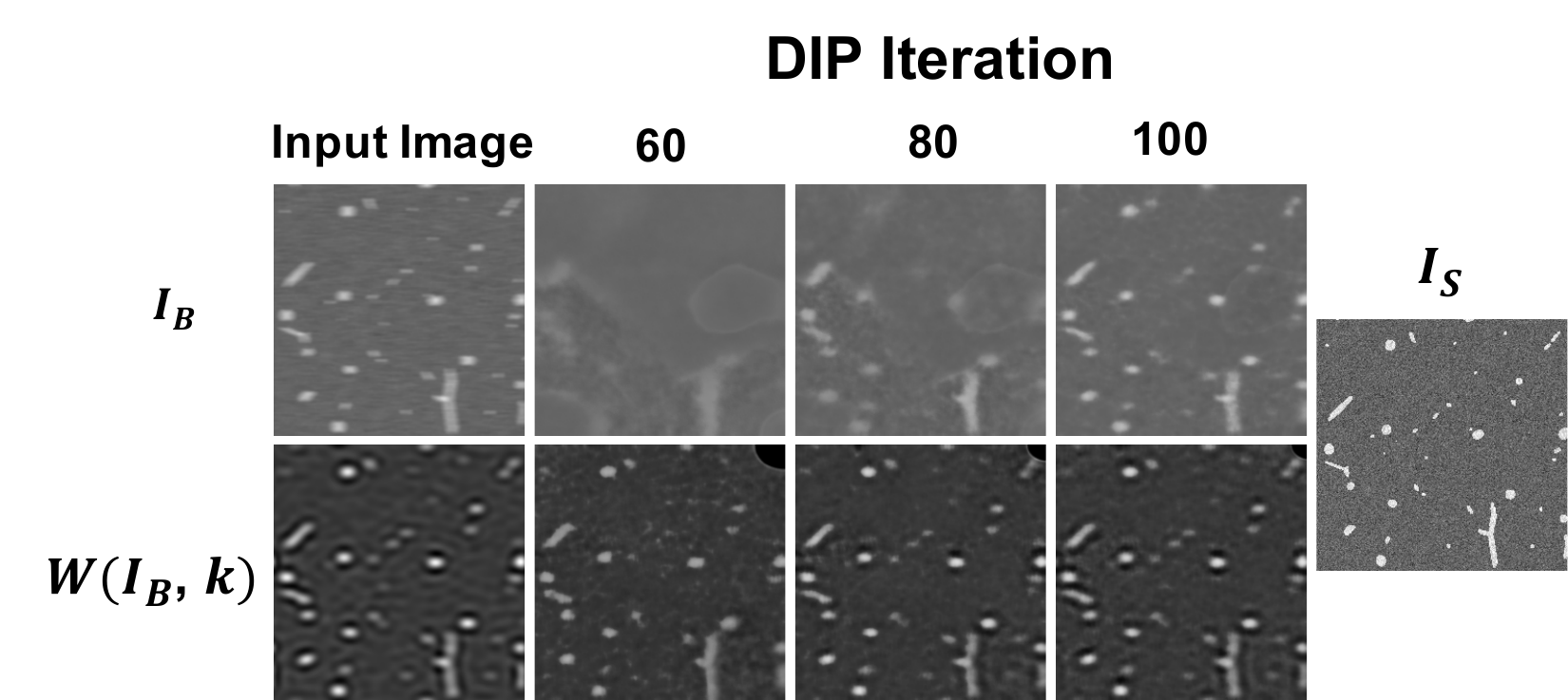}
         \caption{DIP Applied to Blur and Deconvolved Image}
         \label{fig:mot_DIP}
     \end{subfigure}%
     \hspace{0.004\textwidth}
     \begin{subfigure}[b]{0.45\linewidth}
         \includegraphics[width=\linewidth]{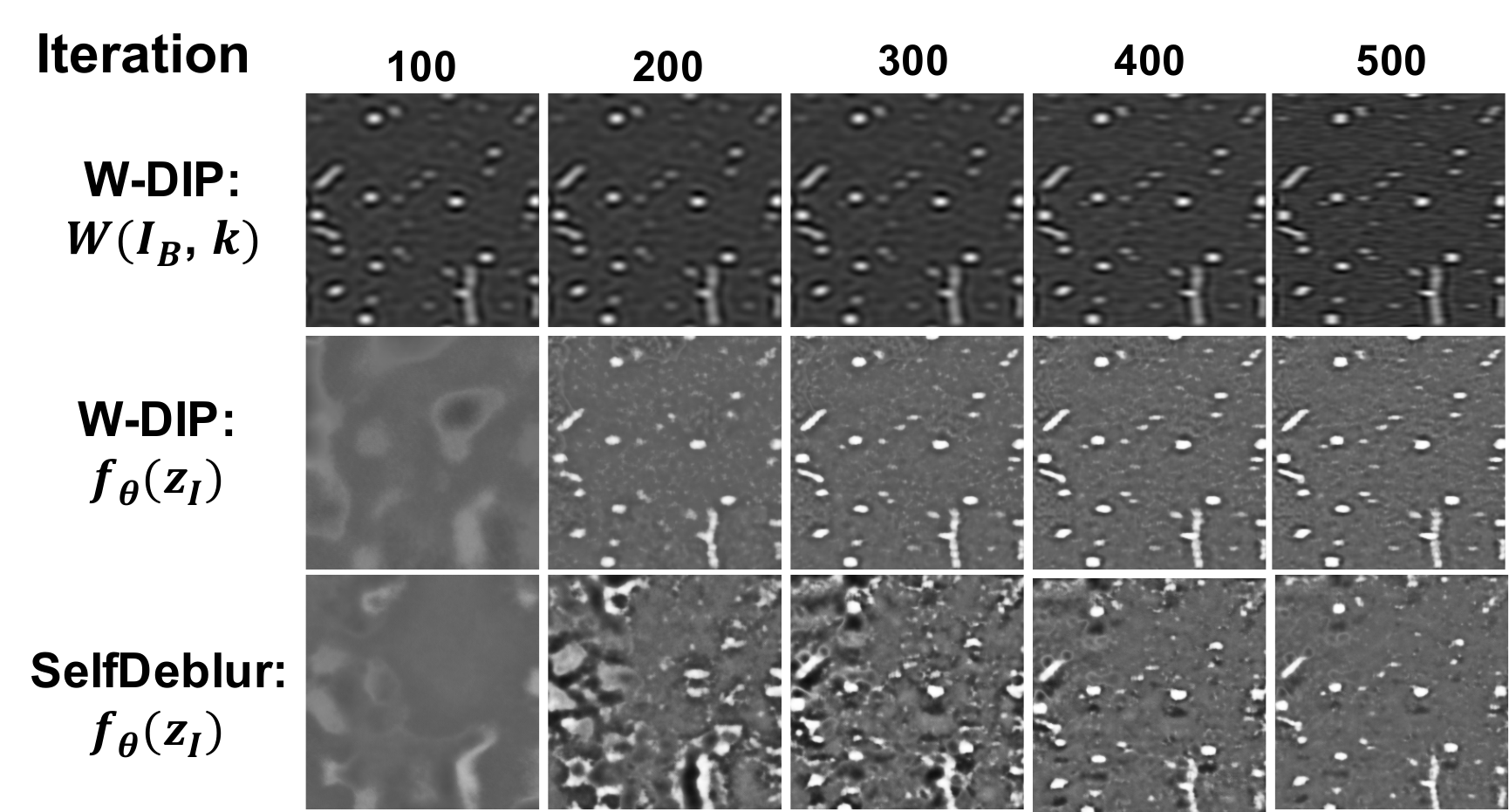}
         \caption{Optimization Comparison Between SelfDeblur and W-DIP}
         \label{fig:mot_comp}
     \end{subfigure}
        \caption{Visual results that motivates incorporating Wiener-deconvolution as a constraint in the DIP framework. (a) Blur image $I_B$ and deconvolved image using Wiener-deconvolution $W(I_B,k)$ with Gaussian kernel are given as input to DIP. We observe that 1) DIP does not generate deconvolution artifacts in the early iterations, 2) DIP learns to generate the deconvolved image much faster compared to the blurry image. These observations make Wiener-deconvolution useful to constrain the image generator network in DIP. (b) Evolution of sharp images during optimization. First and second rows: deconvolved image using Wiener-deconvolution $W(I_B,k)$ and sharp image generated by the image generator network $f_\theta(z_I)$ in the proposed method, last row: sharp image generated by $f_\theta(z_I)$ in SelfDeblur. The results show that proposed method generates sharper images faster with less artifacts compared to SelfDeblur.}
        \label{fig:motivation}
\end{figure*}

Blurred images can occur due to a number of factors ranging from camera motion to the optical set-up. In many cases generating a blur image, $I_B$, can be modeled as the convolution of a sharp image, $I_S$, with a kernel $k$ as shown in Eq. \eqref{eq:general_problem}
\begin{equation}
\begin{split}
    &I_S * k + \eta = I_B,
\end{split}
\label{eq:general_problem}
\end{equation}
where $\eta$ is noise and * is convolution operator.
If $k$ is known then a deconvolution can be performed to recover the original sharp image and this is referred to as non-blind deconvolution. However, often in practice $k$ is not known and needs to be estimated together with $I_S$. The ill-posed nature of the problem requires strong priors on the sharp image and/or kernel to avoid having estimates that do satisfy Eq. \eqref{eq:general_problem}, but are not the sharp version of the given blurred image. One possible formulation of the optimization is Eq. \eqref{eq:general_optimization}.
\begin{equation}
\begin{split}
    &\min_{I_S,k} \|I_S*k - I_B\|^2 + R(I_S) + R(k)
\end{split}
\label{eq:general_optimization}
\end{equation}
where $R(I_S)$ and $R(k)$ are priors on the image and kernel, respectively.
A vast variety of priors have been suggested to regularize the problem in Eq. \eqref{eq:general_problem} in such a way that a solution in a space of interest is found~\cite{krishnan2011blind,cho2009fast,levin2011efficient,xu2013unnatural, zuo2016learning,pan2016l_0}. Common priors include a sparsity constraint on the image gradients and a $L2$-norm on $k$ to avoid the trivial solution. On top or instead of these common priors, recent works suggest more advanced terms such as dark channel prior~\cite{pan2016blind} or edge-based patch priors~\cite{sun2013edge}. Whereas these cleverly constructed priors are powerful they often cause prior-specific artifacts. 

Another approach that has become popular in recent years is learning the mapping from a blurred image to its sharp counter-part. If a large dataset with matching sharp images and their respective blurred version are available in a domain, a network can be trained to learn a successful mapping within the domain~\cite{kupyn2018deblurgan, zheng2019edge,tao2018scale}. There are also methods that combine the learned mappings with classical priors as additional losses~\cite{zhen2019gan}, or learn the priors and incorporate them into classical blind deconvolution frameworks~\cite{li2019blind, asim2020blind}. However, the success of these approaches are often limited to the domain of images they were trained on.

A recent interesting insight by Ulyanov \etal~\cite{ulyanov2018deep} shows that neural network architectures have an intrinsic image generation prior, referred to as ``Deep Image Prior" (DIP), that can be exploited for image restoration tasks. This approach was extended to blind image deconvolution by Ren \etal~\cite{ren2020neural}, SelfDeblur, by having a convolutional neural network (CNN) that generate the estimate for the sharp image and a fully connected network (FCN) that estimates the kernel. By jointly optimizing both networks, impressive blind deconvolution results can be obtained. However, performance of SelfDeblur on a single image fluctuates with different random initialization of network parameters and random noise vectors given as input to the generator networks, which is inherent from DIP. In addition, we observed that SelfDeblur is sensitive to the estimated kernel size, which is a hyper-parameter that should be set prior to optimization and is not always easy to estimate in a blind deconvolution setting.

In previous works, additional regularization methods have been proposed to stabilize and increase restoration performance for inverse problems that utilize DIP. In particular, Romano et al.~\cite{mataev2019deepred} combine DIP with regularization by denoising (RED)~\cite{romano2017little} to improve results on non-blind deconvolution. Another approach adds total variation regularization to DIP and improve performance for denoising~\cite{cascarano2020combining}. Even though these works demonstrate the benefit of the regularization terms, they are not tailored for blind deconvolution. Kotera et al.~\cite{kotera2021improving} specifically focused on improving SelfDeblur by initializing the generator networks in SelfDeblur with the parameters obtained after applying SelfDeblur to the downsampled version of the original image and adding conventional regularization terms. These modifications lead to some performance improvement of SelfDeblur and can also be complementary to our method, but it still suffers from fluctuations. 

\noindent \textbf{Motivation:} One possible regularization that has not been investigated for SelfDeblur to the best of our knowledge, would be to integrate the deconvolution process within the optimization framework. For early iterations the deconvolved image could be a realistic target if the kernel is initialized with a Gaussian and could provide some guidance to the image generator. One simple and effective deconvolution method is Wiener-deconvolution~\cite{wiener1964extrapolation}. It is well known that deconvolution can induce high-frequency artifacts, however DIP has shown to reproduce high frequency features with a lag compared to low frequency features~\cite{ulyanov2018deep}. We investigate if this holds for deconvolution, by reconstructing a deconvolved microscopy image with DIP as is shown in Fig. \ref{fig:mot_DIP} and denoted by $W(I_B,k)$. We compare the reconstruction with that of a blurry image denoted as $I_B$. From the experiment we make two observations. Firstly, as hypothesized the high-frequency artifacts are reproduced with a iteration lag compared to low-frequency features. Secondly, the low-frequency features of the deconvolved image are reproduced much faster for the deconvolved image than for the blurry input image.

\noindent \textbf{Contributions:} Building on these insights, we propose Wiener Guided Deep Image Prior (W-DIP), a method to stabilize the optimization of SelfDeblur by constraining the sharp image generated by the network in SelfDeblur with the sharp image obtained by Wiener-deconvolution~\cite{wiener1964extrapolation}.
To achieve this, we define an auxiliary kernel to apply Wiener-deconvolution to given blurry image and enforce the deconvolved image to be similar to the sharp image generated by the network.
This constraint allows the image generator network to have a realistic sharp target image early on during optimisation since Wiener-deconvolution sharpen edges.
Additionally, the auxiliary kernel is constrained to be similar to the generated kernel to enforce the generated kernel to produce sharp images as well.
We solve the constrained optimization problem using Half Quadratic Splitting (HQS) method and optimize the network parameters and the auxiliary kernel in alternating fashion.
In Fig. \ref{fig:front_figure}, we present a motivating visual example to show the improved deblurring performance achieved by W-DIP compared to the existing methods in the literature. In Fig. \ref{fig:mot_comp}, we show intermediate steps of W-DIP and SelfDeblur on a microscopy image which shows that the proposed method achieves sharper images with less artifacts compared to SelfDeblur.


Our contribution is thus three-fold:

(1) We propose a HQS constrained optimization formulation that integrates Wiener-deconvolution to stabilize and improve the blind image deblurring in the SelfDeblur framework.

(2) We show that, similar to the case of noisy images, high-frequency artifacts introduced by Wiener-deconvolution are only reproduced by DIP with a lag compared to low-frequency features.

(3) We show that initialization of the auxiliary kernel is crucial and propose an initialization strategy that can be automatically adapted to any dataset.

We investigate our contributions by benchmarking with state-of-the-art blind deconvolution algorithms on five different datasets: Levin \cite{levin2009understanding}, Sun \cite{sun2013edge}, Lai \cite{lai2016comparative}, synthetic microscopy\cite{schneider2015joint} as well as a dataset that contains real-life blurry images without a known kernel\cite{lai2016comparative}. In addition, we perform an ablation study to investigate the contribution of each element in our framework.

\begin{figure*}[ht]
     \begin{subfigure}[b]{0.8\linewidth}
         \includegraphics[width=\linewidth]{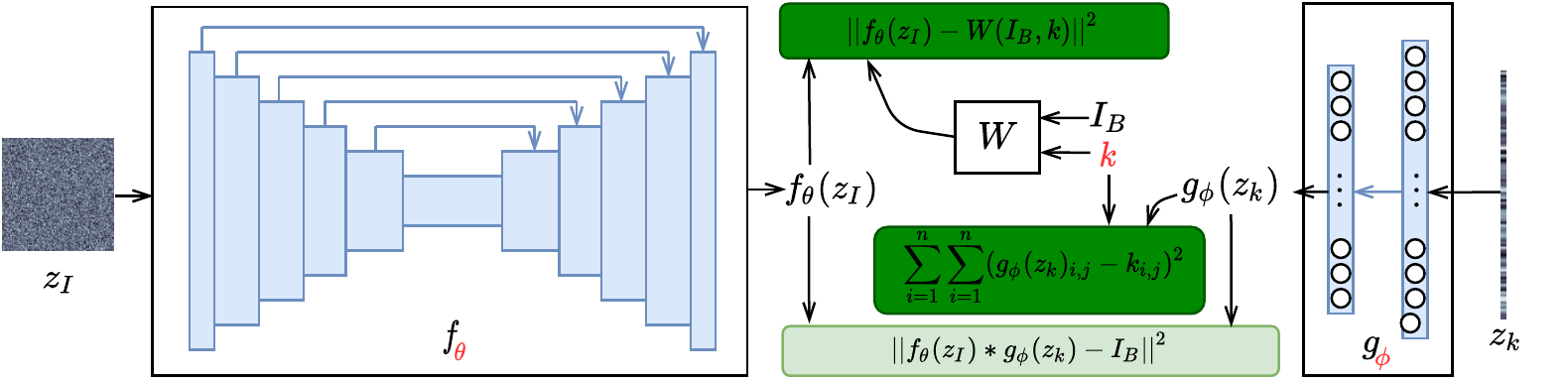}
         \caption{Method Overview}
         \label{fig:dn my_label}
     \end{subfigure}%
     \begin{subfigure}[b]{0.2\linewidth}
         \includegraphics[width=\linewidth]{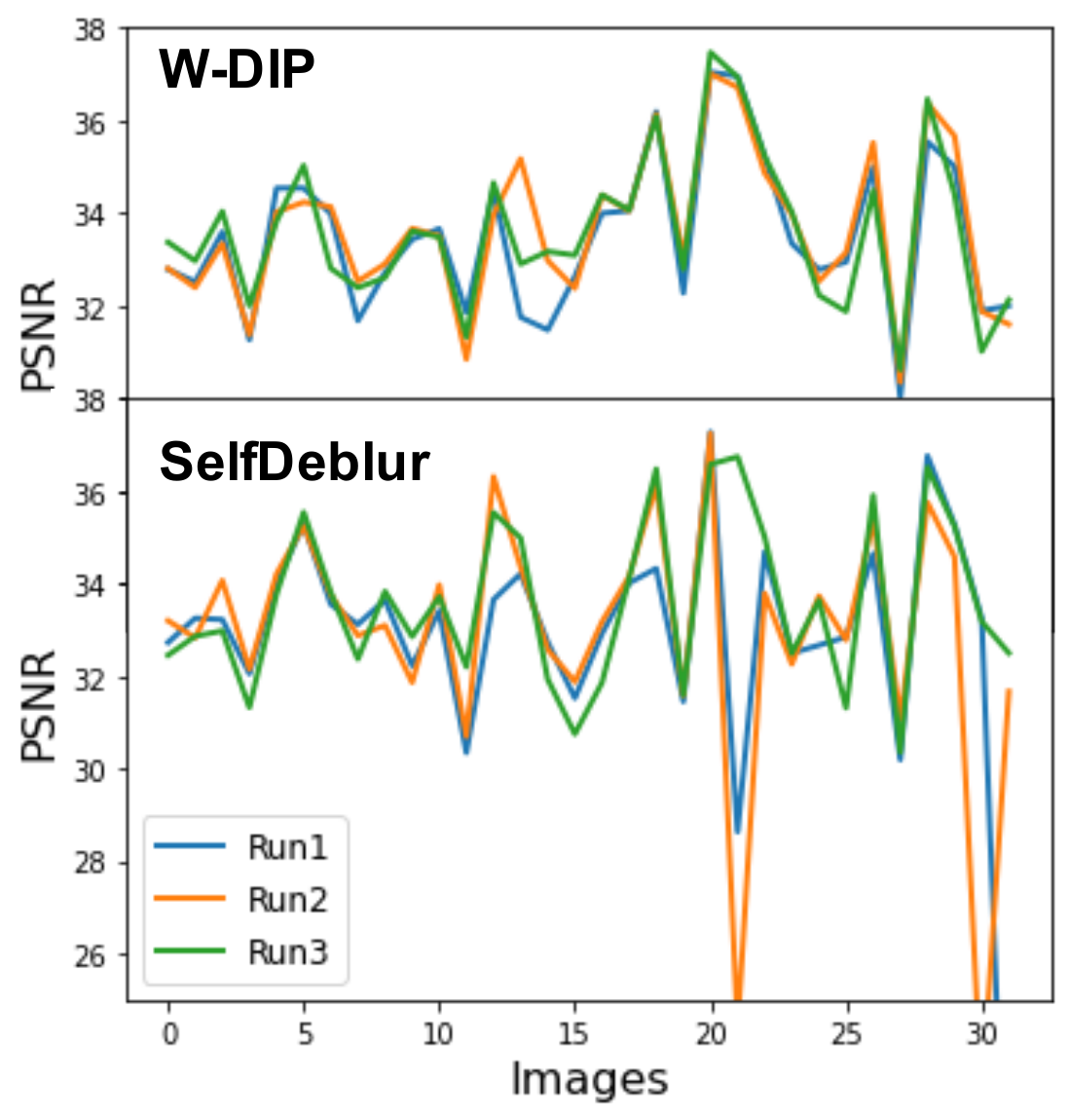}
         \caption{Robustness Comparison}
         \label{fig:in my_label}
     \end{subfigure}
        \caption{a.) Outline of the proposed method W-DIP. The generative network $f_\theta$ generates estimate of sharp image while $g_\phi$ generates kernel estimate. Additionally, an auxiliary kernel, $k$, and blur image are passed through Wiener deconvolution to enforce kernel to generate sharp images. The \textcolor{red}{red} variables are optimised to minimize Eq. \eqref{eq:proposed_unconstrained}. The loss in the light green box correspond to SelfDeblur method and the ones in the darker green boxes indicate the proposed Wiener-deconvolution guided constraints. In b.) both W-DIP and SelfDeblur are run on the Levin \etal~\cite{levin2009understanding} dataset for three runs each. Whereas the performance of SelfDeblur shows a large fluctuation the performance of W-DIP is more stable due to the introduced Wiener-deconvolution guidance.}
        \label{fig:my_label}
\end{figure*}

\section{Method}
\subsection{Background - SelfDeblur}
The use of DIP for blind image deconvolution was proposed by Ren \etal~\cite{ren2020neural} and achieved state of-the-art performance. 
SelfDeblur consists of a CNN $f_{\theta}$ for generating the estimate of the sharp image and FCN $g_{\phi}$ for generating the estimate of the kernel.
The optimization problem to optimize the network parameters $\theta$ and $\phi$ is defined as
\begin{equation}
\begin{split}
    \min_{\theta, \phi} \|f_{\theta}(z_I) * g_{\phi}(z_k) - I_B\|^2 + TV(f_{\theta}(z_I))
\end{split}
\label{eq:selfdeblur}
\end{equation}
in~\cite{ren2020neural} where $z_I$ and $z_k$ are random vectors sampled from $\mathcal{U}(0,1)$.
However, in the published code it has been suggested to remove the TV term and switch from the MSE loss to the structural similarity index measure (SSIM) after 1k iterations which is also confirmed by Kotera \etal in \cite{kotera2021improving}.

\subsection{Proposed method}
In the proposed method, we add constraints to the optimization problem in SelfDeblur given in Eq. \eqref{eq:selfdeblur} to stabilize its training and obtain better blind image deblurring results.
A visualization of the proposed method can be seen in \cref{fig:dn my_label} that shows the different constraints.
As motivated in Sec. 1, we achieve this by constraining the sharp image estimate generated by $f_\theta$ with the deconvolved image obtained after Wiener-deconvolution.
A direct extension of Eq. \eqref{eq:selfdeblur} with such constraint would be
\begin{equation}
\begin{split}
    &\min_{\theta, \phi} \|f_{\theta}(z_I) * g_{\phi}(z_k) - I_B\|^2 \\
    &\textrm{subject to}\; \|f_{\theta}(z_I) - W(I_B, g_\phi(z_k))\|^2=0
\end{split}
\label{eq:proposed_first}
\end{equation}
where $W(I_B, k) = \mathcal{F}^{-1} \{ \frac{|\mathcal{F}\{k\}|^2}{|\mathcal{F}\{k\}|^2 + C} * \frac{ \mathcal{F}\{I_B\}}{\mathcal{F}\{k\} } \}$ and $\mathcal{F}$ is Fourier transform.

The issue with the optimization problem in Eq. \eqref{eq:proposed_first} is that $g_\phi(z_k)$ is similar to uniform kernel in the earlier iterations and Wiener-deconvolution with this kernel does not produce sharp images.
Therefore, we re-write Eq. \eqref{eq:proposed_first} by introducing an auxiliary kernel $k$ with size $n\times n$ and using HQS method as follows:
\begin{equation}
\begin{split}
    &\min_{\theta, \phi, k} \|f_{\theta}(z_I) * g_{\phi}(z_k) - I_B\|^2 \\
    &\textrm{subject to}\; \|f_{\theta}(z_I) - W(I_B, k)\|^2=0\\
    &\;\;\;\;\;\;\;\;\;\;\;\;\;\;\;\;k=g_\phi(z_k)
\end{split}
\label{eq:proposed_second}
\end{equation}
where $k$ is initialized as proposed in the following subsection. We convert the constrained optimization problem using the unconstrained one using the method of Lagrange multipliers
\begin{equation}
\begin{split}
\min_{\theta, \phi, k} &\|f_{\theta}(z_I) * g_{\phi}(z_k) - I_B\|^2 + \alpha \|f_{\theta}(z_I) - W(I_B, k)\|^2 \\ &+\beta \sum_{i=1}^n\sum_{j=1}^{n}\left(g_{\phi}(z_k)_{i,j} - k_{i,j}\right)^2 + \lambda \|k\|^2.
\end{split}
\label{eq:proposed_unconstrained}
\end{equation}
where $\alpha$ and $\beta$ are Lagrange multipliers. Note that, in addition to the constraints in Eq. \eqref{eq:proposed_second}, we add regularization on $k$ with weight $\lambda$. Also, note that we use sum of squared error, not mean square error (MSE) when including the second constraint in Eq. \eqref{eq:proposed_second} to Eq. \eqref{eq:proposed_unconstrained} since MSE is size dependent, i.e. it tends to be lower for larger kernel sizes and higher for the smaller ones since kernels sum up to 1.

Finally, we optimize the network parameters $(\theta, \phi)$ and the auxiliary variable $k$ in an alternating fashion as suggested by HQS to minimize the loss function in Eq. \eqref{eq:proposed_unconstrained}, while keeping $\alpha, \beta$ and $\lambda$ fixed.
The parameters $(\theta, \phi)$ are updated by keeping $k$ fixed and vice versa:
\begin{equation} 
\begin{split}
\min_{\theta, \phi} &\|f_{\theta}(z_I) * g_{\phi}(z_k) - I_B\|^2 + \alpha \|f_{\theta}(z_I)- W(I_B, k)\|^2 \\
&+ \beta \sum_{i=1}^{n}\sum_{i=1}^n (g_{\phi}(z_k)_{i,j} - k_{i,j})^2
\end{split}
\label{eq:outer_optimization}
\end{equation}

\begin{equation}
\begin{split}
\min_{k}\ & \alpha \|f_{\theta}(z_I)- W(I_B, k)\|^2 \\ 
&+ \beta \sum_{i=1}^{n}\sum_{i=1}^n (g_{\phi}(z_k)_{i,j} - k_{i,j})^2 + \lambda \|k\|^2
\end{split}
\label{eq:inner_optimization}
\end{equation}

Note that since the kernel space is sparse in the current formulation, $k$ and $g_{\phi}(z_k)$ could represent the same kernel, but a shifted version to each other. Although both kernel estimates are desirable and satisfy the first and the second terms in Eq. \eqref{eq:proposed_unconstrained}, the kernel matching term weighted by $\beta$ produces large loss due to the translation between the kernels. To circumvent this we align the kernels before computing the kernel matching term. The process is described in more detail in the supplementary.

\subsection{Kernel Initialization} 
The initialization of kernel $k$ is an important parameter, since the starting point of an ill-posed optimization task often restrict the solution space. 
In the case that no prior knowledge of $k$ is available it might be useful to obtain an estimate from the given blurry image $I_B$. 
We propose to achieve this by looking at the power spectral density (PSD) of $I_B$. 
The blurrier an image the more mass will be at the center of the PSD. 

If no prior information is available on the shape of the $k$, we initialize it with a Gaussian and determine the variance of the Gaussian by looking at the PSD of an image as shown in Eqs. \eqref{PSD}. 



\begin{equation} \label{PSD}
  \sum_{i=1}^{Q} \sum_{j=1}^{Q} |\mathcal{F}(y)|^{2}_{i,j} = T,\ G_y = \mathcal{F}^{-1}[\mathcal{N}(0,\,Q^2)]
\end{equation}

\begin{equation} \label{avg_var}
  \sigma_k = \frac{1}{N} \sum_{i=1}^{N} \frac{\sigma(G_{i})}{k_{size_{i}}},\ k_y = \mathcal{N}(0,\,\sigma_k \times k_{size_{y}})
\end{equation}


For each image we calculate the PSD and compute the distance from the center, $Q$, where a certain threshold, $T$, of the mass of the PSD is contained. We then construct a Gaussian in the Fourier Domain with the calculated distance as variance and apply an inverse Fourier transform to the constructed Gaussian. This results in an image-specific $G_y$ that is the size of the image y. To facilitate optimization, by reducing the dimension of the kernel, $G_y$ is cropped to the estimate of the kernel size, however the size of $G_y$ will stay the same in the Fourier domain. To reduce the influence of image specific features an average variance $\sigma_k$ over the dataset consisting of $N$ images is calculated and in addition the variance is normalized by the kernel size estimate $k_{size_i}$ for the respective image as is shown in Eq. \eqref{avg_var}. The initial $k_y$ is then constructed by scaling $\sigma_k$ by the kernel size estimate as shown in Eq. \eqref{avg_var}.

\section{Experimental Results}



\begin{figure*}[ht]
\centering
\captionsetup[subfigure]{labelformat=empty}
     \begin{subfigure}[b]{0.24\linewidth}
         \includegraphics[width=\linewidth]{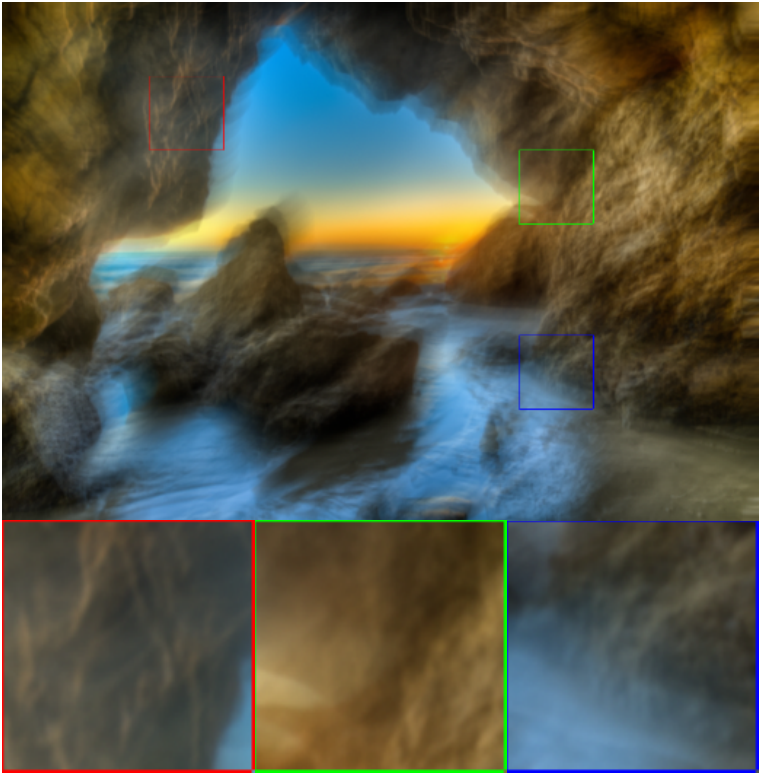}
         \caption{Blurred Image}
         \label{fig:cave_lai_blur}
     \end{subfigure}%
     \hspace{0.008cm}
     \begin{subfigure}[b]{0.24\linewidth}
         \includegraphics[width=\linewidth]{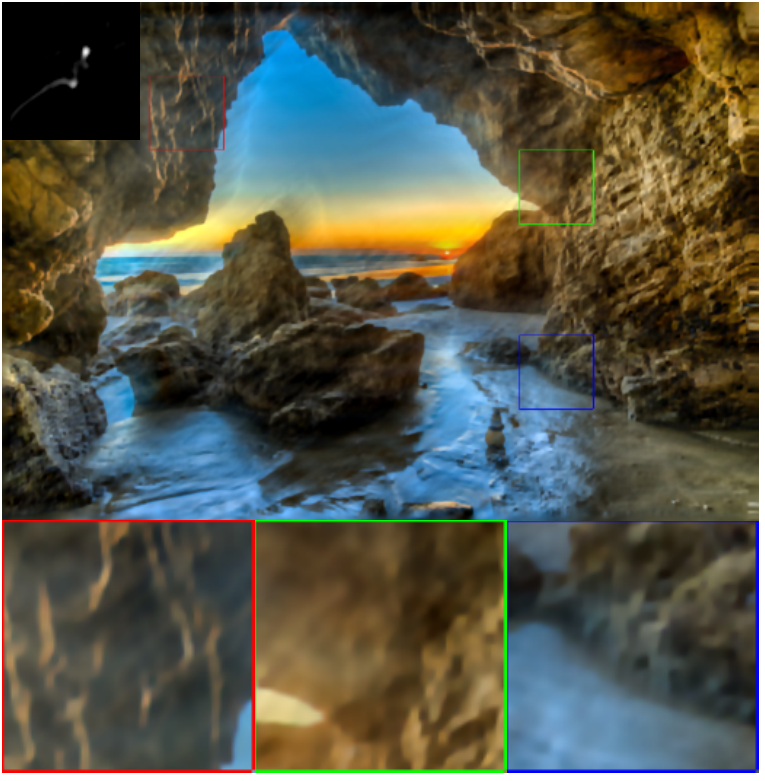}
         \caption{Michaeli \etal~\cite{michaeli2014blind}}
         \label{fig:cave_lai_michaeli}
     \end{subfigure}%
     \hspace{0.008cm}
     \begin{subfigure}[b]{0.24\linewidth}
         \includegraphics[width=\linewidth]{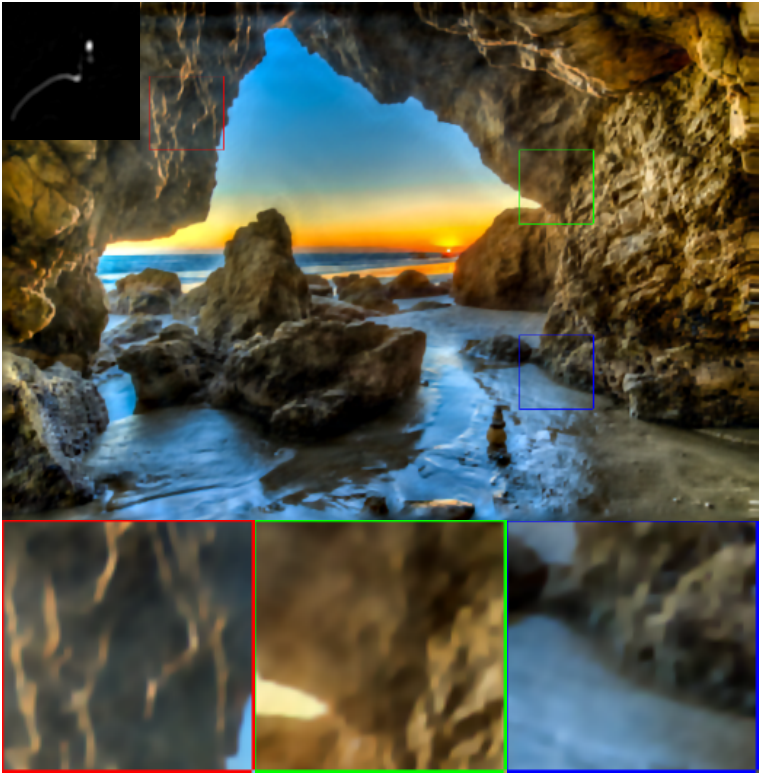}
         \caption{Perrone \etal~\cite{perrone2014total}}
         \label{fig:cave_lai_perrone}
     \end{subfigure}
     \begin{subfigure}[b]{0.24\linewidth}
         \includegraphics[width=\linewidth]{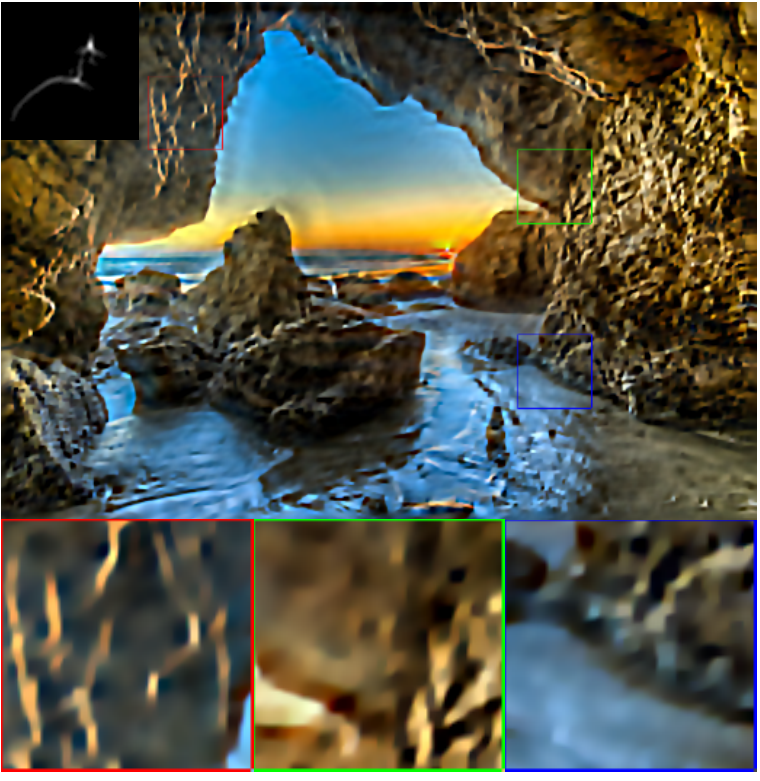}
         \caption{Pan-DCP \etal~\cite{pan2016blind}}
         \label{fig:cave_lai_Pan}
     \end{subfigure}%
  \\
     \begin{subfigure}[b]{0.24\linewidth}
         \includegraphics[width=\linewidth]{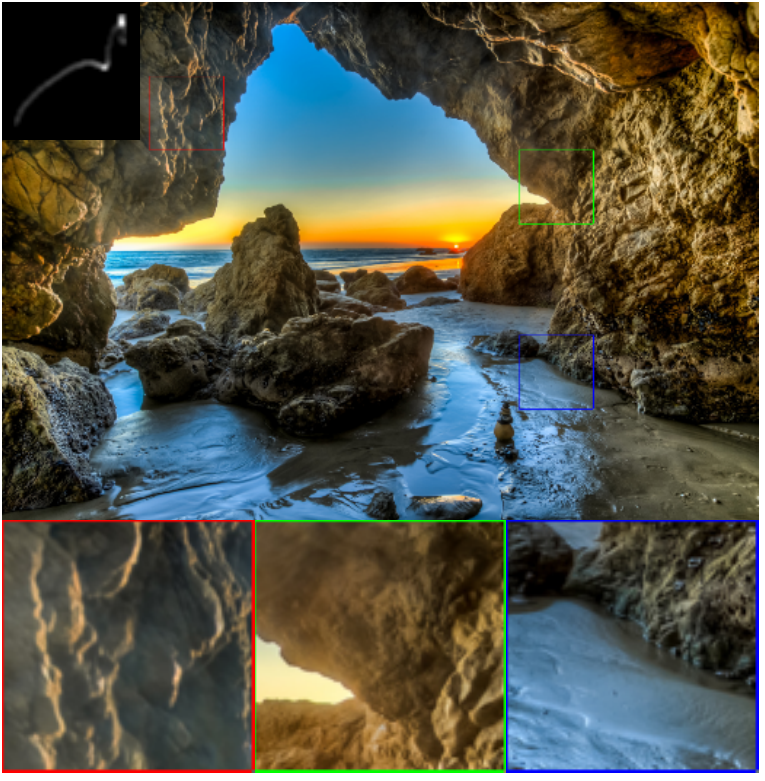}
         \caption{Ground Truth}
         \label{fig:cave_lai_gt}
     \end{subfigure}
     \begin{subfigure}[b]{0.24\linewidth}
         \includegraphics[width=\linewidth]{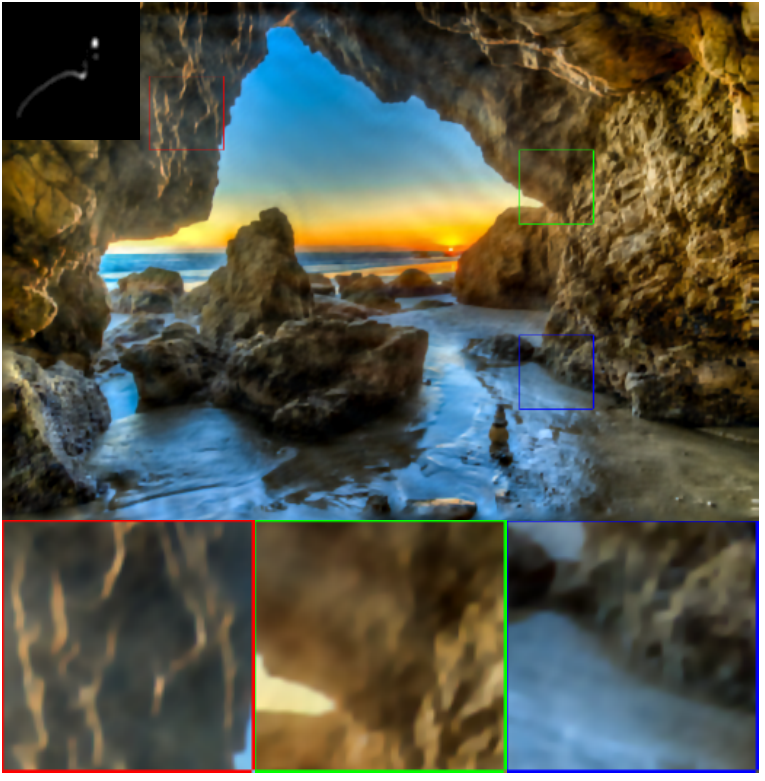}
         \caption{Xu \etal~\cite{xu2013unnatural}}
         \label{fig:cave_lai_xu}
     \end{subfigure}
     \begin{subfigure}[b]{0.24\linewidth}
         \includegraphics[width=\linewidth]{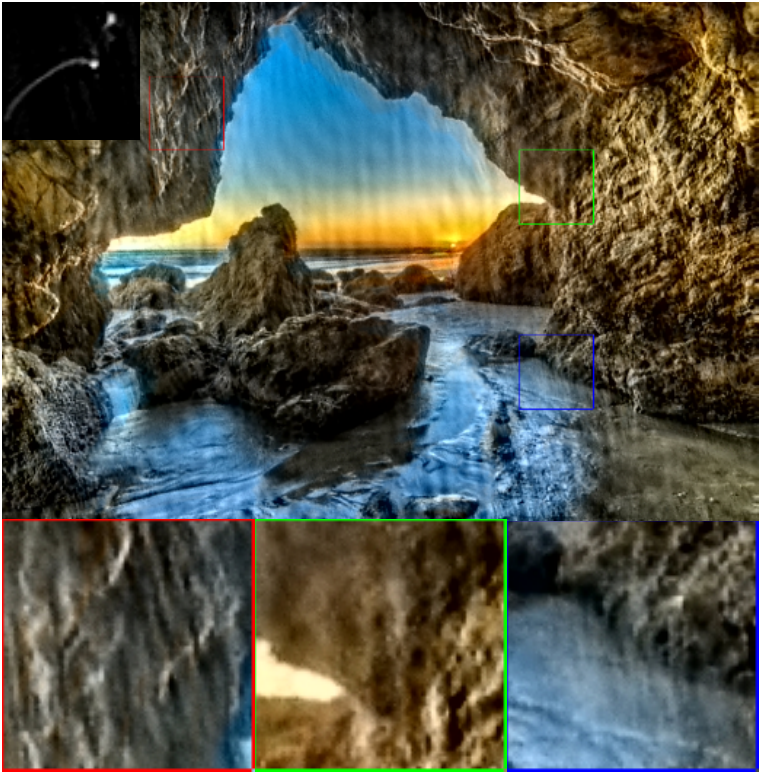}
         \caption{Ren \etal~\cite{ren2020neural}}
         \label{fig:cave_lai_SelfDeblur}
     \end{subfigure}
     \begin{subfigure}[b]{0.24\linewidth}
         \includegraphics[width=\linewidth]{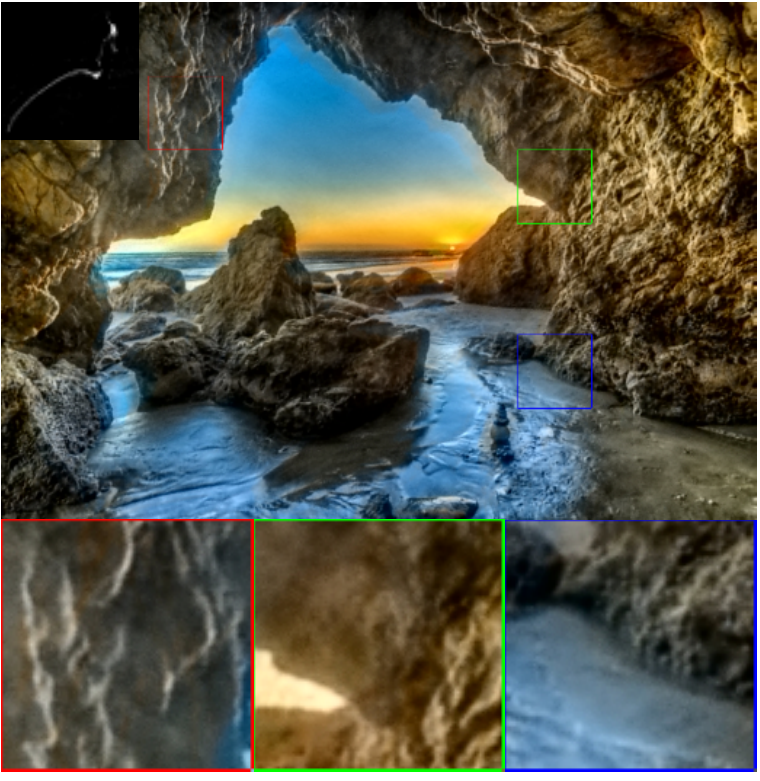}
         \caption{W-DIP}
         \label{fig:cave_lai_WDIP}
     \end{subfigure}
\caption{Visual results from the Lai \etal~\cite{lai2016comparative} dataset to compare W-DIP with five benchmarks. The results of W-DIP is sharper compared to other methods which is especially visible around the rocks.}
\label{fig:lai_example}
\end{figure*}

\textbf{Datasets:} 
We evaluate W-DIP on four widely used natural image datasets and a synthetic microscopy image dataset that we introduced. Levin \etal~\cite{levin2009understanding} introduced a dataset with 4 sharp images that was each convolved with 8 different kernels leading to 32 blurred images. Sun \etal~\cite{sun2013edge} used the same blur kernels, but extended the sharp images from 4 to 80 leading to a total of 640 images. The images are also greyscale as in Levin \etal~\cite{levin2009understanding} but they are significantly larger. Lai \etal~\cite{lai2016comparative} consists of five different image categories, namely: manmade, saturated, text, people and natural. Each category contains five large sharp colour images corresponding to the category name and was blurred with large kernels ranging from $31\times31$ to $75\times75$ in size. This leads to a total of 100 blurred images. In addition, we qualitatively evaluate on real-world blurry images provided by Lai \etal~\cite{lai2016comparative}, where the blurring process is unknown. Lastly, we introduce a dataset of synthetic microscopy images that was blurred with synthetic kernels. Four sharp synthetic images were taken from simulation of Schneider \etal~\cite{schneider2015joint} and cropped to $255\times255$. They were blurred with one kernel from Levin \etal~\cite{levin2009understanding} and three other kernels, that can be found in the supplementary. The kernels are of similar size and shape of the point spread functions expected in two-photon microscopy~\cite{chaigneau2011impact}. In total this leads to 16 blurred images. 

\textbf{Implementation Details:}
The proposed method is implemented in PyTorch~\cite{NEURIPS2019_9015}.
We used Adam~\cite{kingma2014adam} for both optimization steps in alternating optimization each with different learning rates and schedulers.
For the first optimization step in Eq. \eqref{eq:outer_optimization}, we set the initial learning rate to $1e^{-4}$ and decreased it by multiplying with 0.5 at the iterations 2K, 3K, and 4K following the original SelfDeblur implementation. In the second optimization step in Eq. \eqref{eq:inner_optimization}, we used a slightly lower learning rate and a scheduler that depends on the kernel size $n$. 
The reason is that initial estimates of the sharp image generated by $f_\theta$ are worse than that obtained by Wiener-deconvolution since networks are randomly initialized and are given noise as input, thus $\theta$ and $\phi$ should rather adapt to $k$ and $W(I_B,k)$. 
As the estimate from the first optimization gets more reliable, the adaptation speed of $k$ is increased by increasing its learning rate. The larger the estimated kernel size $n$ and thus blur of the image, the longer it takes Eq. \eqref{eq:outer_optimization} to arrive at a reliable solution, thus we include $n$ as a parameter for learning rate adaptation.
We start the second optimization step with the initial learning rate $1e^{-6}$, increase it by a factor of 10 after $70\times(n/10)^{th}$ iteration and two more times after the next two $50\times(n/10)^{th}$ iterations.
The training took place on the in-house GPU cluster mainly consisting of GeForce GTX TITAN X with 12GB memory.

For all evaluations we ran the optimization for 5k iterations, except for the microscopy images where 2k iterations was sufficient to reach peak performance.
Furthermore, we set the weight parameters to: $\alpha=1e^{-3}$, $\beta=1e^{-4}$, $\lambda=1e^{-3}$. 
These parameters were kept constant across all experiments unless otherwise stated and were determined by optimizing performance on four of Levin \etal~\cite{levin2009understanding} images.
We set $C$ in Wiener-deconvolution to a fixed value of 0.025 in all experiments. In addition, for all experiments with SelfDeblur and W-DIP, except on the Sun \etal~\cite{sun2013edge} dataset due to its large size, the experiments were run three times to account for stochasticity. Mean values across the three runs are reported. We do not perform non-blind deconvolution for SelfDeblur and W-DIP as it has been shown that the generated image for SelfDeblur is comparable to that of the non-blind deconvolved version~\cite{ren2020neural}.


\textbf{Evaluation Metrics:}
To evaluate the performance of the deconvolution algorithms we make use of the peak signal-to-noise ratio (PSNR) and structural similarity index meassure (SSIM) between the estimated sharp image and the ground truth. For the Levin \etal~\cite{levin2009understanding} dataset we also include the Error Ratio (E-Ratio) metric ~\cite{levin2011efficient}. In addition, for the microscopy images the vessel structures are of interest from a biological point of view. The structure of these vessels can be obtained by segmentation with Otsu thresholding~\cite{otsu1979threshold} after a preprocessing step where extreme pixel values were removed. The overlap of the segmentation of the ground truth and estimated sharp image was then quantified with the Dice coefficient~\cite{sorensen1948method}.

\subsection{Framework Investigation}
\subsubsection{Ablation Study}

To ensure that each component of the proposed framework is necessary we conduct an ablation study, where the weights for $\alpha$, $\beta$ and $\lambda$ are set to zero individually. In addition, we set the initialization of $k$ to uniform to evaluate the importance of its initialization. We perform the ablation study on a natural image dataset~\cite{levin2009understanding} as well as on the constructed microscopy dataset~\cite{schneider2015joint} as can be seen in \cref{tab:ablation}.

\begin{table}[ht]
\centering
\small
\begin{tabular}{lccc}
\toprule
Method& \begin{tabular}[c]{@{}c@{}}PSNR\end{tabular} &  \begin{tabular}[c]{@{}c@{}}SSIM\end{tabular} &  \begin{tabular}[c]{@{}c@{}}DICE\%\end{tabular} \\ 
\midrule
Levin~\cite{levin2009understanding} & \textbf{33.58$\pm$2.7} & 0.9288$\pm$ 3.9$e^{-4}$ & - \\ 
$\alpha=0$ & 32.74$\pm$8.7 & 0.9086$\pm$4.1$e^{-3}$ & -  \\ 
$\beta=0$ & 33.44$\pm$4.4 & 0.9265$\pm$1.6$e^{-3}$ & -  \\ 
$\lambda=0$ & 33.55$\pm$2.6 & \textbf{0.9298$\pm$3.6$e^{-4}$} & -  \\ 
$k_0=Uni$ & 32.99$\pm$9.3 & 0.9137$\pm$4.9$e^{-3}$ & -  \\ 
\midrule
Microscope & \textbf{20.80$\pm$7.9} & \textbf{0.3977$\pm$2.8$e^{-2}$} & \textbf{76.1$\pm$7.7} \\ 
$\alpha=0$ & 18.97$\pm$18 & 0.3502$\pm$3.6$e^{-2}$ & 65.9$\pm$12  \\ 
$\beta=0$ & 20.5$\pm$9.7 & 0.3833$\pm$2.9$e^{-2}$ & 74.97$\pm$9.2  \\ 
$\lambda=0$ & 20.71$\pm$9.0 & 0.3972$\pm$2.8$e^{-2}$ & 75.86$\pm$8.2  \\ 
$k_0=Uni$ & 20.39$\pm$9.6 & 0.3867$\pm$3.2$e^{-2}$ & 72.16$\pm$9.4  \\ 
\bottomrule
\end{tabular}
\caption{Ablation study of the proposed framework on Levin \etal~\cite{levin2009understanding} and microscopy dataset. To evaluate the contribution of each constraint in our framework we investigate the change in performance when the constraint is left out, by setting the respective weight to zero. The variance is reported to the right of the mean.}
\label{tab:ablation}
\end{table}

It can be seen that the strongest loss in performance in both datasets occurs when the weight $\alpha$, which controls the contribution of the image matching term, is set to zero. This drop empirically demonstrates the importance of the guidance of the Wiener-deconvolution in the proposed method. The performance loss is less pronounced when $\beta$, which controls the contribution of the kernel matching term, is set to zero. Even though it is subtle, the drop is observed for all the metrics, suggesting that this term also plays an important role in the method. For both microscopy and natural images, we see that the influence of the L2-norm on $k$ in the second optimization step is minimal, nevertheless we include the regularization to ensure that the trivial delta function as kernel is avoided. Lastly, we also compare Uniform initialization of $k$ ($k_0 = Uni$) with the proposed initialization and observe that the proposed initialization improves performance in all metrics.


\subsubsection{Robustness Study}

To investigate if the proposed framework can increase the stability of blind deconvolution while using DIP we conducted two experiments. In the first experiment we ran both SelfDeblur and our method for three runs each on the Levin \etal~\cite{levin2009understanding} dataset, utilizing different random seeds in each run.The performance fluctuations for both methods as can be seen in \cref{fig:in my_label}. It can be seen that the performance fluctuation is considerably less for our method compared to SelfDeblur. This is also reflected in the average variance of the PSNR over the three runs. Whereas SelfDeblur has a PSNR variance of 7.18 our method has a variance of 2.71. 


In a second experiment we simulated a realistic scenario where the estimated kernel size used in the optimization is considerably larger than the actual kernel. 
A tight fitting box around the four kernels of the microscopy dataset would be $13\times13$, $35\times35$, $21\times21$ and $27\times27$. Here, we extended these kernels with zero padding to $15\times15$, $41\times41$, $31\times31$ and $37\times37$, respectively. 
In \cref{tab:kernel_size_influence} the performances for SelfDeblur, W-DIP and W-DIP fine-tuned on the microscopy dataset are shown. Fine-tuning was done by optimizing the weight ($\alpha,\beta,\lambda$) of our method on two of the blurry microscopy images and we obtained the weights $\alpha=1e^{-2}$, $\beta=1e^{-1}$, $\lambda=1e^{-2}$. It can be seen when the kernel size estimate is ``good", referring to the first set of smaller kernel sizes, that SelfDeblur performs well, however if the kernel size estimate is too large, ``bad", a performance decrease is observed. For our method the performance decrease in the case of an inaccurate kernel size estimate is significantly less. In addition, fine-tuning the weights ($\alpha,\beta,\lambda$) on two of the blurry images, can further improve the results.

\begin{table}[ht]
\centering
\begin{tabular}{lccc}
\toprule
Method& \begin{tabular}[c]{@{}c@{}}PSNR\end{tabular} &  \begin{tabular}[c]{@{}c@{}}SSIM\end{tabular} &  \begin{tabular}[c]{@{}c@{}}DICE\%\end{tabular} \\ 
\midrule
{\bf SelfDeblur} &  &  &  \\ 
Good $k_{size_{y}}$ & 21.21 & 0.4143 & 83.27  \\ 
Bad $k_{size_{y}}$ & 19.19 & 0.354 & 67.5  \\ 
Difference & \textbf{-2.02} & \textbf{-0.0603} & \textbf{-15.77}  \\ 
\midrule
{\bf W-DIP} &  &  &  \\ 
Good $k_{size_{y}}$ & 21.69 & 0.4141 & 82.9  \\ 
Bad $k_{size_{y}}$ & 20.8 & 0.3977 & 76.1  \\ 
Difference & -0.886 & -0.0164 & -6.8  \\ 
\midrule
{\bf W-DIP Fine-tuned} &  &  &  \\ 
Good $k_{size_{y}}$ & \textbf{22.12} & \textbf{0.4194} & \textbf{84.55}  \\ 
Bad $k_{size_{y}}$ & \textbf{21.84} & \textbf{0.407} & \textbf{82.9}  \\ 
Difference & -0.278 & -0.0124 & -1.65  \\ 
\bottomrule
\end{tabular}
\caption{The performance of SelfDeblur, our method and our method fine-tuned on the microscopy dataset are shown for different kernel size estimates. ``Good" and ``Bad" indicates an accurate and too large kernel size estimate, respectively.}

\label{tab:kernel_size_influence}
\end{table}

\subsection{Quantitative and Qualitative Results}
\subsubsection{Levin \etal~\cite{levin2009understanding} Dataset}

We evaluate our algorithm on the Levin \etal~\cite{levin2009understanding} dataset and compare it against other state-of-the-art unsupervised blind deconvolution methods~\cite{krishnan2011blind,cho2009fast,levin2011efficient,xu2013unnatural,sun2013edge, zuo2016learning,pan2016blind}, whose results are provided by Ren \etal~\cite{ren2020neural}. SelfDeblur was trained with the exact same learning rate and scheduler as we used for the networks in the constrained optimization and was in accordance with the code provided by the authors. The results can be seen in \cref{tab:LevinQuant}.


\begin{figure}[ht]
\centering
\captionsetup[subfigure]{labelformat=empty}
     \begin{subfigure}[b]{0.40\linewidth}
         \includegraphics[width=\linewidth]{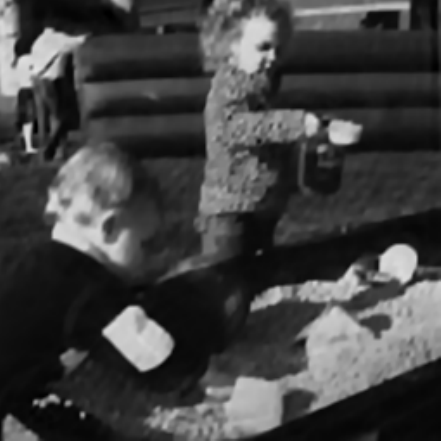}
         \caption{Pan \etal~\cite{pan2016blind}, SSIM:0.9350}
         \label{fig:ssim_pan}
     \end{subfigure}%
     \hspace{0.001\textwidth}
     \begin{subfigure}[b]{0.40\linewidth}
         \includegraphics[width=\linewidth]{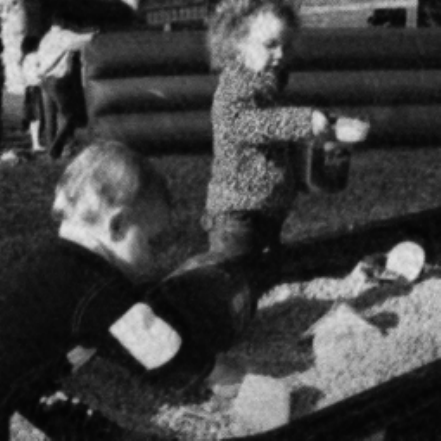}
         \caption{W-DIP, SSIM:0.9180}
         \label{fig:ssim_wdip}
     \end{subfigure}
        \caption{Visual example of SSIM favoring over-smoothing of image on Levin \etal~\cite{levin2009understanding} dataset}
        \label{fig:ssim}
\end{figure}

\begin{table}[ht]
\centering
\begin{tabular}{lcccc}
\toprule
Method& \begin{tabular}[c]{@{}c@{}}PSNR\end{tabular} &  \begin{tabular}[c]{@{}c@{}}SSIM\end{tabular} & 
\begin{tabular}[c]{@{}c@{}}E-Ratio\end{tabular} & \begin{tabular}[c]{@{}c@{}}Time\end{tabular} \\ 
\midrule
k known & 35.42 & 0.9576 & 1.000 & - \\ 
Cho\&Lee~\cite{cho2009fast}* & 30.29 & 0.8973 & 1.860 & 1.395 \\ 
Levin \etal~\cite{levin2011efficient}* & 30.90 & 0.9173 & 1.759 & 78.26 \\ 
Krishnan \etal~\cite{krishnan2011blind}* & 29.87 & 0.8680 & 2.549 & 8.940 \\ 
Xu \etal~\cite{xu2013unnatural}* & 31.27 & 0.9148 & 1.608 & 1.184  \\ 
Zuo \etal~\cite{zuo2016learning}* & 32.42 & 0.9353 & 1.334 & 11.00 \\ 
Pan \etal~\cite{pan2016blind}* & 32.88 & \textbf{0.9386} & 1.232 & 295.2 \\
Sun \etal~\cite{sun2013edge}* & 33.10 & 0.9385 & 1.363 & 191.0 \\
SelfDeblur~\cite{ren2020neural} & 33.01 & 0.9134 & 1.207 & 220.0 \\
\midrule
W-DIP & \textbf{33.50} & 0.9270 & \textbf{1.087} & 280.6  \\ 
\bottomrule
\end{tabular}
\caption{Quantative evaluation on Levin \etal~\cite{levin2009understanding} dataset, while comparing to other state-of-the art blind deconvolution approaches. The asterisk next to a method indicates that non-blind deconvolution~\cite{levin2011efficient} was performed with the found kernel.}
\label{tab:LevinQuant}
\end{table}

Compared to other blind deconvolution methods W-DIP outperforms the benchmarks both in PSNR and especially with regard to Error Ratio, however comes at a higher computational cost. Since the second optimization step only requires optimization over a small parameter space, the computational cost is only marginally larger than that of SelfDeblur. The lower SSIM score compared with other benchmarks such as Pan \etal~\cite{pan2016blind} can be explained by the SSIM metric that favors over-smoothed images as seen in \cref{fig:ssim}. Whereas the strong smoothing might be beneficial for small images, we see in the Lai \etal~\cite{lai2016comparative} dataset that it is not beneficial for more challenging datasets, where we outperform all benchmarks also with regard to the SSIM metric.

\begin{figure*}[ht]
\centering
\captionsetup[subfigure]{labelformat=empty}
     \begin{subfigure}[b]{0.18\linewidth}
         \includegraphics[width=\linewidth]{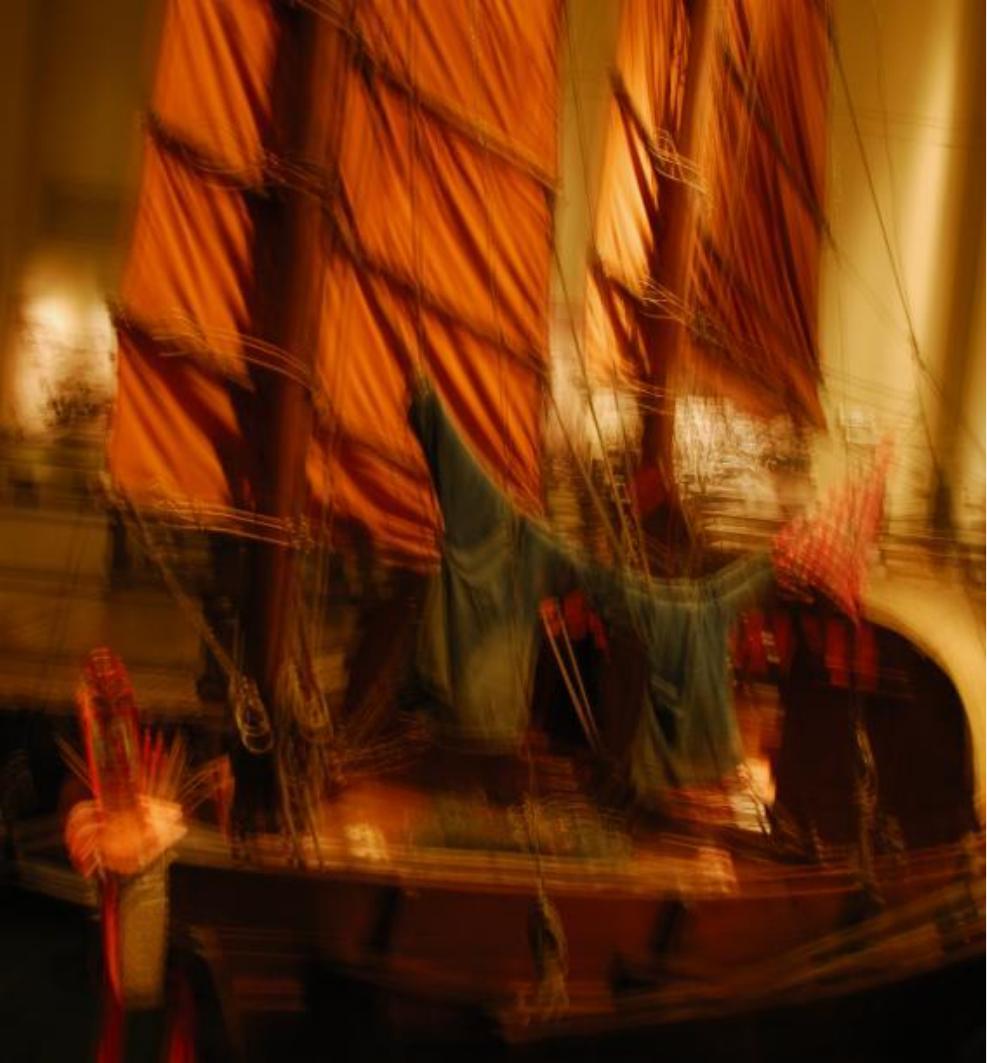}
         \caption{Blurred Image}
         \label{fig:boat_blur}
     \end{subfigure}
     \begin{subfigure}[b]{0.18\linewidth}
         \includegraphics[width=\linewidth]{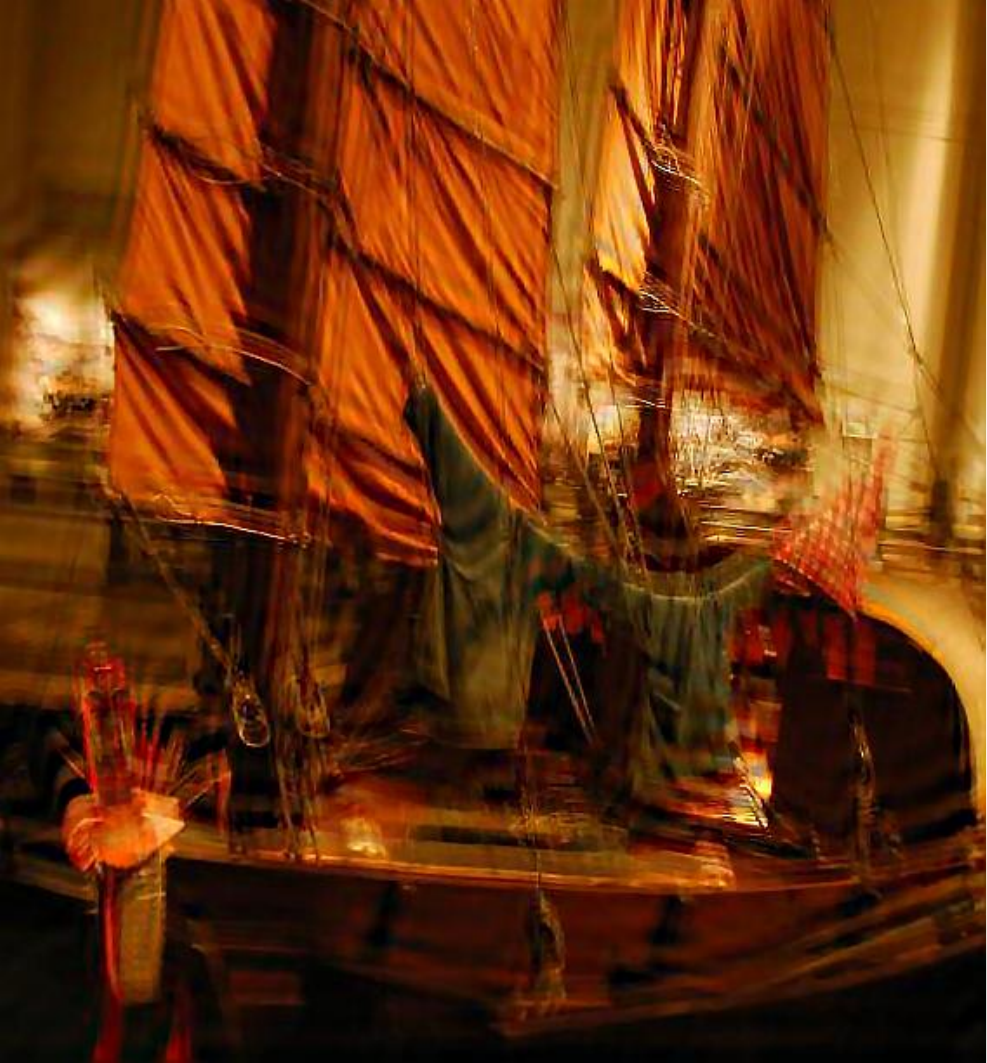}
         \caption{Michaeli \etal~\cite{michaeli2014blind}}
         \label{fig:real_michaeli}
     \end{subfigure}
     \begin{subfigure}[b]{0.18\linewidth}
         \includegraphics[width=\linewidth]{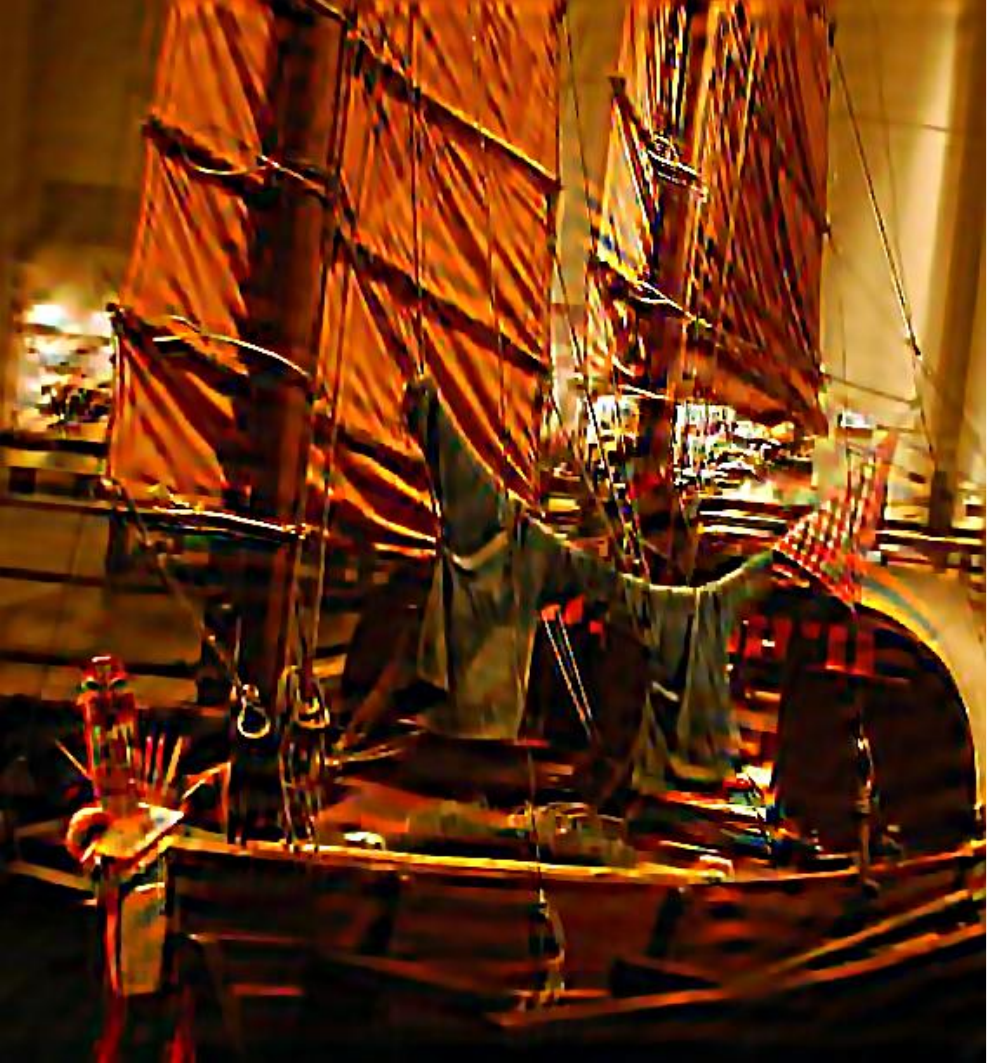}
         \caption{Pan-DCP \etal~\cite{pan2016blind}}
         \label{fig:real_pan}
     \end{subfigure}
     \begin{subfigure}[b]{0.18\linewidth}
         \includegraphics[width=\linewidth]{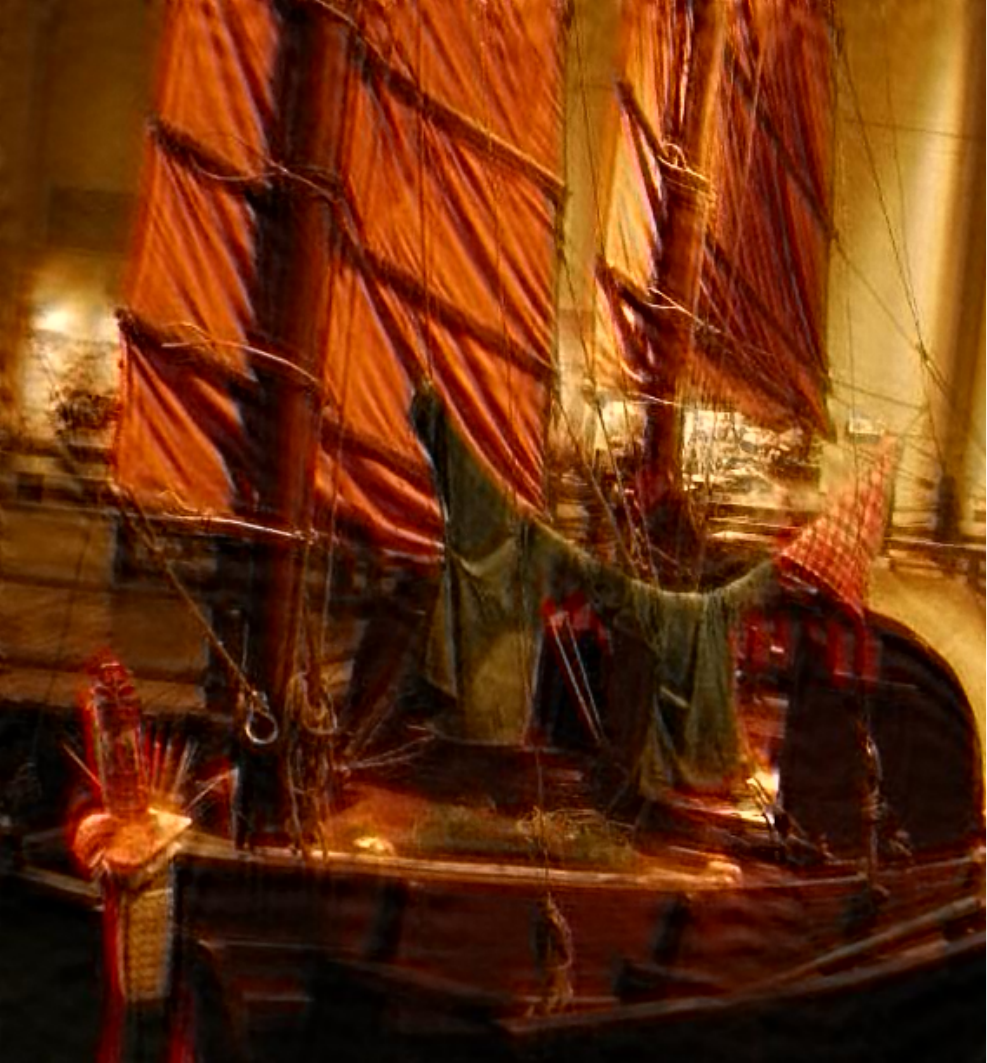}
         \caption{Ren \etal~\cite{ren2020neural}}
         \label{fig:real_ren}
     \end{subfigure}
     \begin{subfigure}[b]{0.18\linewidth}
         \includegraphics[width=\linewidth]{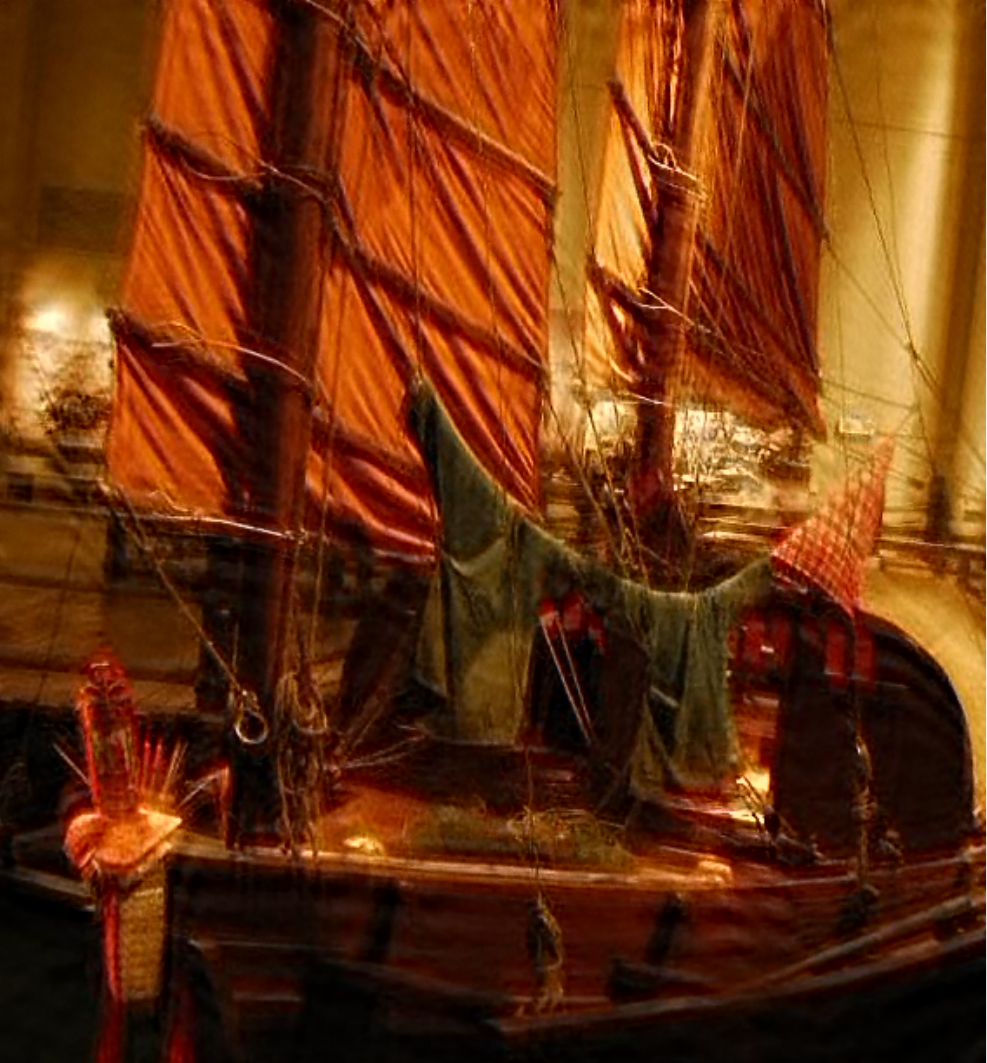}
         \caption{W-DIP}
         \label{fig:real_wdip}
     \end{subfigure}
        \caption{Real-world blur example from the Lai \etal~\cite{lai2016comparative} dataset comparing the generated image of our method with three benchmarks}
        \label{fig:real_world}
\end{figure*}
\begin{table*}[t]
\centering
\begin{tabular}{lcccccc}
\toprule
Method& \begin{tabular}[c]{@{}c@{}}Manmade\end{tabular} &  \begin{tabular}[c]{@{}c@{}}Saturated\end{tabular} & 
\begin{tabular}[c]{@{}c@{}}Text\end{tabular} &
\begin{tabular}[c]{@{}c@{}}People\end{tabular} &
\begin{tabular}[c]{@{}c@{}}Natural\end{tabular} &
\begin{tabular}[c]{@{}c@{}}Average\end{tabular} \\ 
\midrule
Cho\&Lee~\cite{cho2009fast}* & 16.35/0.3890 & 14.05/0.4927 & 14.87/0.4429 & 19.90/0.5560 & 20.14/0.5198 & 17.06/0.4801\\ 
Xu\&Jia ~\cite{xu2010two}* & 19.23/0.6540 & 14.79/0.5632 & 18.56/0.7171 & 25.32/\underline{0.8517} & \underline{23.03}/\underline{0.7542} & 20.18/0.7080 \\ 
Xu \etal~\cite{xu2013unnatural}* & 17.99/0.5986 & 14.53/0.5383 & 17.64/0.6677 & 14.40/0.8133 & 21.58/0.6788 & 19.23/0.6593 \\ 
Michaeli \etal~\cite{xu2013unnatural}* & 17.43/0.4189 & 14.14/0.4914 & 16.23/0.4686 & 23.35/0.6999 & 20.70/0.5116 & 18.37/0.5181  \\ 
Perrone \etal~\cite{perrone2014total}* & 17.41/0.550 & 14.24/0.5107 & 16.94/0.592 & 22.77/0.7347 & 21.04/0.676 & 18.48/0.6130 \\ 
Pan-DCP \etal~\cite{pan2016blind}* & 18.59/0.5942 & 16.52/0.6322 & 17.42/0.6193 & 24.03/0.7719 & 22.60/0.6984 & 19.89/0.6656 \\
Pan-L0 \etal~\cite{pan2016l_0}* & 16.92/0.5316 & 14.62/0.5451 & 16.87/0.6030  & 23.36/0.7822 & 20.92/0.6622 & 18.54/0.6248 \\
SelfDeblur~\cite{ren2020neural} & 19.39/\textbf{0.6660} & \underline{18.55}/\underline{0.7066} & \textbf{23.21}/\underline{0.7754} & \underline{27.49}/0.8172 & 21.81/0.6585 & 22.09/0.7247 \\
\midrule
W-DIP & \textbf{19.51}/\underline{0.6653} & \textbf{18.66}/\textbf{0.7217} & \underline{22.48}/\textbf{0.7832} & \textbf{28.35}/\textbf{0.8586} & \textbf{23.47}/\textbf{0.7562} & \textbf{22.49}/\textbf{0.7570}  \\ 
\bottomrule
\end{tabular}
\caption{Quantitative evaluation on Lai \etal~\cite{lai2016comparative} dataset. Performance of the algorithms are shown per image category with the PSNR value left and SSIM value right of the slash, respectively. The asterisk next to a method indicates that non-blind deconvolution~\cite{krishnan2009fast,whyte2014deblurring} was performed with the found kernel. Bold and underlined metrics are the best and second best result, respectively.}
\label{tab:LaiQuant}
\end{table*}
\subsubsection{Lai \etal~\cite{lai2016comparative} Dataset}
Whereas the Levin \etal~\cite{levin2009understanding} dataset is limited to 32 images and relatively small kernels ranging from $13\times13$ to $27\times27$ as well as small greyscale images, $255\times255$, the Lai \etal~\cite{lai2016comparative} dataset contains large kernels and large colour images. To deal with colour images we utilize the same approach as Ren \etal~\cite{ren2020neural} by splitting the images into the YCbCr channels respectively and only optimize for the Y channel. In addition, the dataset is split into distinct image categories, which we also inspect individually as can be seen in \cref{tab:LaiQuant}. We compare against other state-of-the-art blind deconvolution benchmarks~\cite{cho2009fast, xu2010two, xu2013unnatural, michaeli2014blind, perrone2014total, pan2016l_0, pan2016blind} and again use the results for the benchmarks that was reported and shared by Ren \etal~\cite{ren2020neural}. 

Whereas other benchmarks such as Xu and Jia~\cite{xu2010two} perform exceptionally well for a specific category, such as ``Natural" in this case, but show a weak performance in another category for example ``Saturated", our method produce state-of-the art results in every category and thus also leads to the best average performance when compared over all image categories. Especially in the categories ``People" and ``Natural" our method show a clear improvement over SelfDeblur. One example image can be seen in \cref{fig:lai_example}. The figure shows that our method creates the least amount of artifacts for the given image.

\subsubsection{Sun \etal~\cite{sun2013edge}}
We also perform experiments on a large dataset by Sun \etal~\cite{sun2013edge} containing 640 large images. The results shows that W-DIP is more stable, by producing a lower variance across all images, while also achieving a higher mean PSNR and SSIM as shown in \cref{tab:sun}. A qualitative comparison can be found in the supplementary.


\begin{table}[ht]
\centering
\begin{tabular}{lccc}
\toprule
Method& \begin{tabular}[c]{@{}c@{}}PSNR\end{tabular} &  \begin{tabular}[c]{@{}c@{}}SSIM\end{tabular}\\ 
\midrule
SelfDeblur~\cite{ren2020neural} & 28.8$\pm$7.4 & 0.7383$\pm$8.6$e^{-3}$  \\ 
W-DIP & \textbf{29.07 $\pm$ 6.7} & \textbf{0.7465$\pm$6.3$e^{-3}$}  \\ 
\bottomrule
\end{tabular}
\caption{Here we present the results of blind deconvolution on the Sun \etal~\cite{sun2013edge} dataset. We report the mean PSNR and SSIM values with theeir respective variance on the right side.}
\label{tab:sun}
\end{table}


\subsubsection{Real-World Images With Unknown Blur}
For real-world application the blurring operation is often not known. More complex blurring operations such as non-uniform blurring can occur. To investigate the performance of W-DIP on these challenging images we compare the performance with benchmarks~\cite{michaeli2014blind, pan2016blind,ren2020neural} in \cref{fig:real_world}. It can be seen that our result is close to that of SelfDeblur, while outperforming other benchmarks on this image. More qualitative comparisons can be found in the supplementary.

\section{Conclusion and Limitations}

\textbf{Conclusion:} In this paper, we proposed an unsupervised blind image deconvolution method by guiding optimization of DIP with Wiener-deconvolution to obtain more stable and enhanced deblurring performance.
We performed experiments on five different datasets and ablation studies have been used to validate the proposed components of our framework. 
The results show that W-DIP improves existing methods in terms of both image quality and training stability. 

\textbf{Limitations: } Despite the fact that we achieve significant improvement in terms of stability, sensitivity of DIP to random initialisation still remain a problem. 
In addition, in this work we did not quantitatively investigate the performance on non-uniform blurred images. This could be a possible challenge for the current method, since uniform blurring operations are used in the data-fitting term and for deconvolution. Future work could try and extend the approach to different blur operations and investigate using more powerful deconvolution methods.

{\small
\bibliographystyle{ieee_fullname}
\bibliography{egbib}

\begin{thebibliography}{10}\itemsep=-1pt

\bibitem{asim2020blind}
Muhammad Asim, Fahad Shamshad, and Ali Ahmed.
\newblock Blind image deconvolution using deep generative priors.
\newblock {\em IEEE Transactions on Computational Imaging}, 6:1493--1506, 2020.

\bibitem{cascarano2020combining}
Pasquale Cascarano, Andrea Sebastiani, Maria~Colomba Comes, Giorgia Franchini,
  and Federica Porta.
\newblock Combining weighted total variation and deep image prior for natural
  and medical image restoration via admm.
\newblock {\em arXiv preprint arXiv:2009.11380}, 2020.

\bibitem{chaigneau2011impact}
Emmanuelle Chaigneau, Amanda~J Wright, Simon~P Poland, John~M Girkin, and
  R~Angus Silver.
\newblock Impact of wavefront distortion and scattering on 2-photon microscopy
  in mammalian brain tissue.
\newblock {\em Optics express}, 19(23):22755--22774, 2011.

\bibitem{cho2009fast}
Sunghyun Cho and Seungyong Lee.
\newblock Fast motion deblurring.
\newblock In {\em ACM SIGGRAPH Asia 2009 papers}, pages 1--8, 2009.

\bibitem{kingma2014adam}
Diederik~P Kingma and Jimmy Ba.
\newblock Adam: A method for stochastic optimization.
\newblock {\em arXiv preprint arXiv:1412.6980}, 2014.

\bibitem{kotera2021improving}
Jan Kotera, Filip {\v{S}}roubek, and V{\'a}clav {\v{S}}m$\acute{\iota}$dl.
\newblock Improving neural blind deconvolution.
\newblock In {\em 2021 IEEE International Conference on Image Processing
  (ICIP)}, pages 1954--1958. IEEE, 2021.

\bibitem{krishnan2009fast}
Dilip Krishnan and Rob Fergus.
\newblock Fast image deconvolution using hyper-laplacian priors.
\newblock {\em Advances in neural information processing systems},
  22:1033--1041, 2009.

\bibitem{krishnan2011blind}
Dilip Krishnan, Terence Tay, and Rob Fergus.
\newblock Blind deconvolution using a normalized sparsity measure.
\newblock In {\em CVPR 2011}, pages 233--240. IEEE, 2011.

\bibitem{kupyn2018deblurgan}
Orest Kupyn, Volodymyr Budzan, Mykola Mykhailych, Dmytro Mishkin, and
  Ji{\v{r}}{\'\i} Matas.
\newblock Deblurgan: Blind motion deblurring using conditional adversarial
  networks.
\newblock In {\em Proceedings of the IEEE conference on computer vision and
  pattern recognition}, pages 8183--8192, 2018.

\bibitem{lai2016comparative}
Wei-Sheng Lai, Jia-Bin Huang, Zhe Hu, Narendra Ahuja, and Ming-Hsuan Yang.
\newblock A comparative study for single image blind deblurring.
\newblock In {\em Proceedings of the IEEE Conference on Computer Vision and
  Pattern Recognition}, pages 1701--1709, 2016.

\bibitem{levin2009understanding}
Anat Levin, Yair Weiss, Fredo Durand, and William~T Freeman.
\newblock Understanding and evaluating blind deconvolution algorithms.
\newblock In {\em 2009 IEEE Conference on Computer Vision and Pattern
  Recognition}, pages 1964--1971. IEEE, 2009.

\bibitem{levin2011efficient}
Anat Levin, Yair Weiss, Fredo Durand, and William~T Freeman.
\newblock Efficient marginal likelihood optimization in blind deconvolution.
\newblock In {\em CVPR 2011}, pages 2657--2664. IEEE, 2011.

\bibitem{li2019blind}
Lerenhan Li, Jinshan Pan, Wei-Sheng Lai, Changxin Gao, Nong Sang, and
  Ming-Hsuan Yang.
\newblock Blind image deblurring via deep discriminative priors.
\newblock {\em International journal of computer vision}, 127(8):1025--1043,
  2019.

\bibitem{mataev2019deepred}
Gary Mataev, Peyman Milanfar, and Michael Elad.
\newblock Deepred: Deep image prior powered by red.
\newblock In {\em Proceedings of the IEEE/CVF International Conference on
  Computer Vision Workshops}, pages 0--0, 2019.

\bibitem{michaeli2014blind}
Tomer Michaeli and Michal Irani.
\newblock Blind deblurring using internal patch recurrence.
\newblock In {\em European conference on computer vision}, pages 783--798.
  Springer, 2014.

\bibitem{otsu1979threshold}
Nobuyuki Otsu.
\newblock A threshold selection method from gray-level histograms.
\newblock {\em IEEE transactions on systems, man, and cybernetics},
  9(1):62--66, 1979.

\bibitem{pan2016l_0}
Jinshan Pan, Zhe Hu, Zhixun Su, and Ming-Hsuan Yang.
\newblock $ l\_0 $-regularized intensity and gradient prior for deblurring text
  images and beyond.
\newblock {\em IEEE transactions on pattern analysis and machine intelligence},
  39(2):342--355, 2016.

\bibitem{pan2016blind}
Jinshan Pan, Deqing Sun, Hanspeter Pfister, and Ming-Hsuan Yang.
\newblock Blind image deblurring using dark channel prior.
\newblock In {\em Proceedings of the IEEE Conference on Computer Vision and
  Pattern Recognition}, pages 1628--1636, 2016.

\bibitem{NEURIPS2019_9015}
Adam Paszke, Sam Gross, Francisco Massa, Adam Lerer, James Bradbury, Gregory
  Chanan, Trevor Killeen, Zeming Lin, Natalia Gimelshein, Luca Antiga, Alban
  Desmaison, Andreas Kopf, Edward Yang, Zachary DeVito, Martin Raison, Alykhan
  Tejani, Sasank Chilamkurthy, Benoit Steiner, Lu Fang, Junjie Bai, and Soumith
  Chintala.
\newblock Pytorch: An imperative style, high-performance deep learning library.
\newblock In H. Wallach, H. Larochelle, A. Beygelzimer, F. d\textquotesingle
  Alch\'{e}-Buc, E. Fox, and R. Garnett, editors, {\em Advances in Neural
  Information Processing Systems 32}, pages 8024--8035. Curran Associates,
  Inc., 2019.

\bibitem{perrone2014total}
Daniele Perrone and Paolo Favaro.
\newblock Total variation blind deconvolution: The devil is in the details.
\newblock In {\em Proceedings of the IEEE Conference on Computer Vision and
  Pattern Recognition}, pages 2909--2916, 2014.

\bibitem{ren2020neural}
Dongwei Ren, Kai Zhang, Qilong Wang, Qinghua Hu, and Wangmeng Zuo.
\newblock Neural blind deconvolution using deep priors.
\newblock In {\em Proceedings of the IEEE/CVF Conference on Computer Vision and
  Pattern Recognition}, pages 3341--3350, 2020.

\bibitem{romano2017little}
Yaniv Romano, Michael Elad, and Peyman Milanfar.
\newblock The little engine that could: Regularization by denoising (red).
\newblock {\em SIAM Journal on Imaging Sciences}, 10(4):1804--1844, 2017.

\bibitem{schneider2015joint}
Matthias Schneider, Sven Hirsch, Bruno Weber, G{\'a}bor Sz{\'e}kely, and
  Bjoern~H Menze.
\newblock Joint 3-d vessel segmentation and centerline extraction using oblique
  hough forests with steerable filters.
\newblock {\em Medical image analysis}, 19(1):220--249, 2015.

\bibitem{sorensen1948method}
Th~A Sorensen.
\newblock A method of establishing groups of equal amplitude in plant sociology
  based on similarity of species content and its application to analyses of the
  vegetation on danish commons.
\newblock {\em Biol. Skar.}, 5:1--34, 1948.

\bibitem{sun2013edge}
Libin Sun, Sunghyun Cho, Jue Wang, and James Hays.
\newblock Edge-based blur kernel estimation using patch priors.
\newblock In {\em IEEE International Conference on Computational Photography
  (ICCP)}, pages 1--8. IEEE, 2013.

\bibitem{tao2018scale}
Xin Tao, Hongyun Gao, Xiaoyong Shen, Jue Wang, and Jiaya Jia.
\newblock Scale-recurrent network for deep image deblurring.
\newblock In {\em Proceedings of the IEEE Conference on Computer Vision and
  Pattern Recognition}, pages 8174--8182, 2018.

\bibitem{ulyanov2018deep}
Dmitry Ulyanov, Andrea Vedaldi, and Victor Lempitsky.
\newblock Deep image prior.
\newblock In {\em Proceedings of the IEEE conference on computer vision and
  pattern recognition}, pages 9446--9454, 2018.

\bibitem{whyte2014deblurring}
Oliver Whyte, Josef Sivic, and Andrew Zisserman.
\newblock Deblurring shaken and partially saturated images.
\newblock {\em International journal of computer vision}, 110(2):185--201,
  2014.

\bibitem{wiener1964extrapolation}
Norbert Wiener et~al.
\newblock {\em Extrapolation, interpolation, and smoothing of stationary time
  series: with engineering applications}, volume~8.
\newblock MIT press Cambridge, MA, 1964.

\bibitem{xu2010two}
Li Xu and Jiaya Jia.
\newblock Two-phase kernel estimation for robust motion deblurring.
\newblock In {\em European conference on computer vision}, pages 157--170.
  Springer, 2010.

\bibitem{xu2013unnatural}
Li Xu, Shicheng Zheng, and Jiaya Jia.
\newblock Unnatural l0 sparse representation for natural image deblurring.
\newblock In {\em Proceedings of the IEEE conference on computer vision and
  pattern recognition}, pages 1107--1114, 2013.

\bibitem{zhen2019gan}
Ada Zhen, Robert~L Stevenson, et~al.
\newblock Gan based image deblurring using dark channel prior.
\newblock {\em Electronic Imaging}, 2019(13):136--1, 2019.

\bibitem{zheng2019edge}
Shuai Zheng, Zhenfeng Zhu, Jian Cheng, Yandong Guo, and Yao Zhao.
\newblock Edge heuristic gan for non-uniform blind deblurring.
\newblock {\em IEEE Signal Processing Letters}, 26(10):1546--1550, 2019.

\bibitem{zuo2016learning}
Wangmeng Zuo, Dongwei Ren, David Zhang, Shuhang Gu, and Lei Zhang.
\newblock Learning iteration-wise generalized shrinkage--thresholding operators
  for blind deconvolution.
\newblock {\em IEEE Transactions on Image Processing}, 25(4):1751--1764, 2016.

\end{thebibliography}
}

\end{document}